\documentclass[10pt,twocolumn,letterpaper]{article}

\usepackage{cvpr}              
\definecolor{cvprblue}{rgb}{0.21,0.49,0.74}
\usepackage[pagebackref,breaklinks,colorlinks,allcolors=cvprblue]{hyperref}

\usepackage[T1]{fontenc}
\usepackage{bbding}
\usepackage{microtype}
\usepackage{graphicx}
\usepackage{pgfplots}
\usepackage[utf8]{inputenc} %
\usepackage[T1]{fontenc}    %
\usepackage{nicefrac}       %
\usepackage{microtype}      %
\usepackage{amsmath}
\usepackage{amssymb}
\usepackage{wrapfig}
\usepackage{amsfonts}
\usepackage{mathtools}
\usepackage{amsthm}
\usepackage{multirow}
\usepackage{booktabs}
\usepackage{amsfonts,pifont,tabularx}
\usepackage{algorithm}
\usepackage{algpseudocode}
\usepackage{multirow}
\usepackage{tabu}
\usepackage{tabularx}
\usepackage{textcomp}
\usepackage{xcolor}
\usepackage{url}
\usepackage{mdframed}
\usepackage{multirow, makecell}
\PassOptionsToPackage{table,xcdraw}{xcolor}
\usepackage{colortbl}
\usepackage{soul}
\usepackage{bm}
\usepackage{latexsym}
\usepackage{mathtools}
\usepackage{enumitem}
\usepackage{diagbox}
\usepackage{wrapfig}
\usepackage{tikz}
\usepackage{pgfplots}
\usepgfplotslibrary{groupplots}
\pgfplotsset{compat=newest}
\usepackage{array}
\newcolumntype{?}{!{\vrule width 1pt}}
\usepackage{caption} 
\usepackage{wrapfig}

\usepackage{soul}
\usepackage[normalem]{ulem}
\usepackage{xspace}
\usepackage{lipsum}
\usepackage{etoc}
\usepackage{caption} 
\usepackage{cuted} 
\usepackage{subcaption} %
\setlist[itemize]{leftmargin=1.5em}
\usepackage[most]{tcolorbox}
\usepackage[table]{xcolor}
\newtcolorbox{promptbox}[1]{
  title={#1}, breakable,
  colback=gray!12, colframe=black!35,
  boxrule=0.4pt, arc=2pt,
  left=6pt, right=6pt, top=6pt, bottom=6pt,
  listing only,
  listing options={
    basicstyle=\ttfamily\footnotesize,
    breaklines=true,
    showstringspaces=false
  }
}
\newcommand{\todoc}[2]{{\textcolor{#1}{\textbf{#2}}}}

\newcommand{\todored}[1]{{\todoc{red}{\textbf{[[#1]]}}}}

\newcommand{\ma}[1]{\todored{Ma: #1}}

\newtheorem{theorem}{Theorem}[section]          

\theoremstyle{remark}

\makeatletter
\newcommand{\suppstart}[1]{%
  \clearpage
  \onecolumn
  \thispagestyle{plain}%
  \setcounter{page}{1}%
  \begin{center}
    {\Large\bfseries #1\par}%
    \vspace{0.5em}
    \vspace{1.0em}
  \end{center}%
  \setcounter{section}{0}%
  \renewcommand{\thesection}{\Alph{section}}%
  \renewcommand{\thesubsection}{\thesection.\arabic{subsection}}%
}
\makeatother

\def\paperID{1427} 
\def\confName{CVPR}
\def\confYear{2026}

\newcommand{\sys}
{\mbox{\textsc{PromptMiner}}\xspace}

\title{\sys: Black-Box Prompt Stealing against Text-to-Image Generative Models via Reinforcement Learning and Fuzz Optimization}

\author{%
   \textbf{Mingzhe Li}\textsuperscript{1},
   \textbf{Renhao Zhang}\textsuperscript{1},
   \textbf{Zhiyang Wen}\textsuperscript{1},
   \textbf{Siqi Pan}\textsuperscript{2}\\
   \textbf{Bruno Castro da Silva}\textsuperscript{1},
   \textbf{Juan Zhai}\textsuperscript{1},
    \textbf{Shiqing Ma}\textsuperscript{1}\\
   \textsuperscript{1}University of Massachusetts, Amherst 
   \textsuperscript{2}Dolby Laboratories\\
}

\begin{document}
\maketitle

\begin{abstract}
Text-to-image (T2I) generative models such as Stable Diffusion and FLUX can synthesize realistic, high-quality images directly from textual prompts. 
The resulting image quality depends critically on well-crafted prompts that specify both subjects and stylistic modifiers, which have become valuable digital assets. 
However, the rising value and ubiquity of high-quality prompts expose them to security and intellectual-property risks. 
One key threat is the \textbf{prompt stealing attack}, i.e., the task of recovering the textual prompt that generated a given image. 
Prompt stealing enables unauthorized extraction and reuse of carefully engineered prompts, yet it can also support beneficial applications such as data attribution, model provenance analysis, and watermarking validation. 
Existing approaches often assume white-box gradient access, require large-scale labeled datasets for supervised training, or rely solely on captioning without explicit optimization, limiting their practicality and adaptability. 
To address these challenges, we propose \textbf{\sys}, a black-box prompt stealing framework that decouples the task into two phases: 
(1) a reinforcement learning–based optimization phase to reconstruct the primary subject, 
and (2) a fuzzing-driven search phase to recover stylistic modifiers. 
Experiments across multiple datasets and diffusion backbones demonstrate that \sys achieves superior results, with CLIP similarity up to 0.958 and textual alignment with SBERT up to 0.751, surpassing all baselines. 
Even when applied to in-the-wild images with unknown generators, it outperforms the strongest baseline by 7.5\% in CLIP similarity, demonstrating better generalization. 
Finally, \sys maintains strong performance under defensive perturbations, highlighting remarkable robustness.
Code: \url{https://github.com/aaFrostnova/PromptMiner}
\end{abstract}
\section{Introduction}
\label{sec:intro}
\begin{figure}[t]
  \centering
  \includegraphics[width=\linewidth]{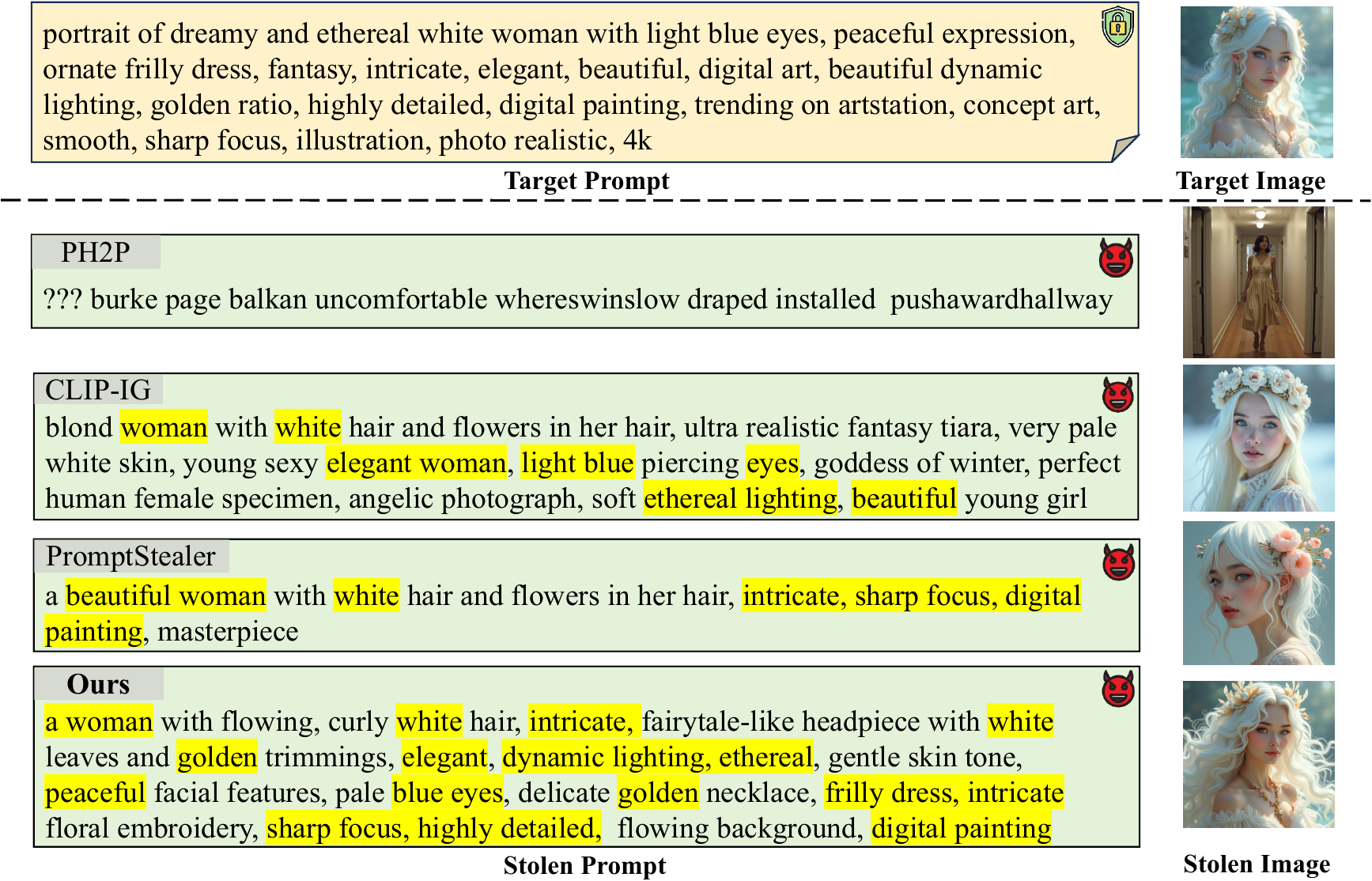}
  \caption{Illustration of the prompt stealing. Given a target image, the attacker aims to recover the prompt to generate a similar image.}
  \label{fig:illustration}
\end{figure}
Text-to-image (T2I) generative models such as \textit{Stable Diffusion}~\cite{Rombach_2022_CVPR} \textit{FLUX}~\cite{flux2024}, and \textit{DALL·E}~\cite{ramesh2022hierarchical} have become widely used in our daily lives, due to their excellent performance in synthesizing realistic and creative images from descriptive prompts. 
A key factor in successfully generating high-quality images is careful prompt design. 
Well-crafted prompts often encode sensitive details and typically combine precise \textit{subject} (core visual entity) with rich\textit{ modifiers} (stylistic and compositional cues)~\cite{liu2023designguidelinespromptengineering, hao2023optimizing}. 
However, obtaining high-quality prompts is far from trivial. 
Crafting an effective prompt requires substantial domain knowledge and iterative experimentation, as minor changes in phrasing, token order, or modifier choice can dramatically alter the generated image's composition. 
Moreover, such well-engineered prompts have become valuable digital assets—carefully guarded by creators and even traded on specialized marketplaces such as PromptBase~\cite{promptbase} and PromptHero~\cite{prompthero}, highlighting the growing economic and creative significance of prompt design in the generative AI ecosystem. 

The growing value and ubiquity of high-quality prompts expose them to security and intellectual-property threats, most notably prompt stealing attacks or prompt inversion methods~\cite{shen2024promptstealingattackstexttoimage,ph2p2024cvpr,kim2025visuallyguideddecodinggradientfree, li2025editor}. 
In such attacks, adversaries aim to recover the most precise descriptive prompt to generate the target image as closely as possible. 
Beyond infringing intellectual property, an attacker can resell the stolen prompt on third-party marketplaces, directly monetizing others' work. 
This practice erodes the commercial interests of both platforms and original creators. 
At the same time, reconstructing prompts from generated images also serves as a beneficial forensic tool for trustworthy AI, enabling model provenance analysis and watermark validation~\citep{wang2024did, wang2023diagnosis, naseh2024iteratively}.

Currently, there is a wide array of cutting-edge prompt stealing or prompt inversion methods. 
First, white-box prompt inversion methods~\cite{10.5555/3666122.3668341, ph2p2024cvpr} mostly focus on the subject and often yield linguistically awkward prompts. 
Crucially, the white-box assumption is rarely valid in commercial deployments, limiting practicality. 
In contrast, black-box methods like BLIP~\cite{li2022blip} and VGD~\cite{kim2025visuallyguideddecodinggradientfree} achieve prompt recovery by combining existing components such as CLIP, image captioning models, or large language models. 
However, they similarly lack the ability to reconstruct modifiers. 
CLIP-IG~\cite{clip_interrogator_github} and PromptStealer~\cite{shen2024promptstealingattackstexttoimage} consider both the subject and modifiers when recovering prompts.
However, they lack an explicit optimization process, relying instead on direct generation from pretrained models. This direct generation often lacks similarity in image generation. 
Specifically, PromptStealer trains an inversion model for subject generation and a classification model for modifier prediction on large-scale, labeled datasets (i.e., Lexica~\cite{shen2024promptstealingattackstexttoimage}). 
This dataset-specific training also introduces a strong risk of overfitting, limiting the model's generalization ability across diverse images.

To overcome these limitations, we propose \sys, a novel and effective black-box prompt stealing framework that achieves both semantic precision and stylistic fidelity without requiring model gradients or training on large-scale, labeled datasets. 
The core insight is to decouple prompt stealing into (i) a reinforcement learning-based optimization phase to reconstruct the primary subject, and (ii) a fuzzing-driven search phase to recover stylistic modifiers. 
Unlike prior white-box methods that depend on gradient access, \sys\ first leverages reinforcement learning~(RL)~\cite{10.5555/3312046} to perform black-box optimization, precisely capturing the core visual entities and yielding concise, content-faithful subjects.
Specifically, we formulate the prompt inversion as a Markov decision process (MDP) and solve it using RL with a shaped reward for faster convergence.
Then, to enrich the recovered prompt with modifiers, we introduce a fuzz testing-powered optimization phase that treats modifier discovery as controlled exploration in prompt space: carefully designed mutators propose targeted edits, while a capacity-constrained seed pool maintains and exploits the most promising candidates.
Together, these two phases enable efficient discrete prompt space search, preserving semantic grounding while systematically injecting modifiers crucial for high-fidelity image synthesis. As shown in \autoref{fig:illustration}, \sys outperforms baseline methods by producing more accurate prompts and generating images most similar to the target.


We conduct a comprehensive experimental assessment of \sys. 
Our results on the three widely-used datasets (e.g., MS COCO~\cite{10.1007/978-3-319-10602-1_48}, Flickr~\cite{young2014image}, and Lexica~\cite{SQBZ24}) under four state-of-the-art T2I generation model (e.g., Stable Diffusion v1.5~\citep{Rombach_2022_CVPR}, SDXL-Turbo~\cite{sauer2023adversarialdiffusiondistillation}, FLUX.1 dev~\cite{flux2024}, and Stable Diffusion 3.5 Medium~\cite{esser2024scaling}) show that \sys is more effective that baselines with CLIP similarity up to 0.958 and textual alignment with SBERT up to 0.751. 
Moreover, \sys also outperforms the strongest baseline by 7.5\% in CLIP similarity on \textit{in-the-wild} images where the target generative models are unknown, demonstrating its strong generalization ability. 
We further evaluate the robustness of \sys under several potential defense strategies, including random noise injection, patch perturbation, and textual watermarking. 
These post-processing defenses exert only marginal influence on our performance, indicating that \sys is inherently resilient to low-level visual perturbations due to its semantics-driven optimization process. 
Our contributions are summarized as follows:
\begin{itemize}
\item We introduce \sys, a novel and effective black-box prompt stealing framework against text-to-image generative models. 
Unlike existing methods that assume white-box gradient access, rely on direct caption generation without explicit optimization, or depend on large, labeled datasets with poor adaptability, \sys efficiently refines prompts, delivering precise and expressive reconstructions without relying on model gradients or large-scale labeled datasets.
\item We formulate prompt stealing as an MDP and optimize an adapter on top of a frozen captioner with reinforcement learning, using potential-based reward shaping for dense guidance and a fuzz testing-powered optimization phase to discover effective modifiers. 
This design yields concise, precise subjects and systematically injects style and composition cues.
\item We conduct extensive evaluations across three widely-used datasets and four advanced text-to-image generative models, where \sys\ achieves state-of-the-art performance in both image similarity and textual alignment. 
\sys\ further generalizes to \textit{in-the-wild} images (unknown target models) and remains robust under common post-processing defenses.
\end{itemize}
\section{Related Work}
\label{sec:related_work}


\noindent \textbf{Image Captioning.} A straightforward way to obtain prompts is to apply captioning models or advanced vision language models (VLMs), such as BLIP~\cite{li2022blip}, BLIP-2~\cite{li2023blip}, or GPT-4o~\cite{hurst2024gpt}. Given an input image $x$, a captioning model or a VLM generates a textual description as the prompt in an autoregressive manner:
\begin{equation}
P(p_{0:T}\!\mid\!x) = \prod_{t=0}^{T} P(p_t \!\mid\! p_{<t}, x),
\end{equation}
where $p_t$ denotes the $t$-th token, $T$ is the sequence length determined when the model emits an end-of-sequence token, and $p_{0:T}=\{p_0,\ldots,p_T\}$ represents the complete prompt. 
At each step $t$, the model conditions on both the visual features of $x$ and the textual prefix $p_{<t}$ to predict the distribution over the next token. While these models generate fluent descriptions, they fail to ensure high similarity when re-generating the image with a diffusion model. 

\noindent \textbf{Prompt Stealing Attack and Prompt Inversion.} Recent gradient-based works~\cite{10.5555/3666122.3668341, ph2p2024cvpr, li2025editor} focus on optimizing textual prompts to better align with image features. PEZ~\cite{10.5555/3666122.3668341} introduces a gradient-based discrete optimization method for hard prompt discovery with CLIP~\cite{mokady2021clipcap}, and PH2P~\cite{ph2p2024cvpr} proposes hard prompt optimization that aligns prompts with image features directly in diffusion models. However, these inverted prompts often lack semantic fluency and naturalness, resulting in outputs unintelligible to humans. Another recent work, Visually Guided Decoding (VGD)~\cite{kim2025visuallyguideddecodinggradientfree}, is gradient-free and uses a large language model plus CLIP-based guidance to generate coherent, human-readable prompts. Nevertheless, VGD primarily focuses on identifying a single salient subject and overlooks the richer modifier details. 
Building on findings from~\cite{liu2023designguidelinespromptengineering}, effective prompts typically follow a \textit{“Subject, Modifiers”} structure, where modifiers capture nuanced visual semantics beyond the main entity. In contrast, CLIP Interrogator~\cite{clip_interrogator_github} and PromptStealer~\cite{shen2024promptstealingattackstexttoimage} extend the generation process by explicitly adding descriptive modifiers from a predefined pool or by decomposing prompts into subject and modifier components. 
However, both approaches lack a targeted optimization process that refines the composed prompts toward the specific image. 
Moreover, PromptStealer relies on large-scale labeled datasets for supervised training, which limits its scalability and adaptability to unseen domains. 
Similarly, Prometheus~\cite{zhao2025effectivepromptstealingattack} depends heavily on an extensive prompt base constructed from pre-collected prompt data to guide its modifier search, making it less flexible in the open world.

\begin{figure*}[t] 
  \centering
  \includegraphics[width=1.0\linewidth]{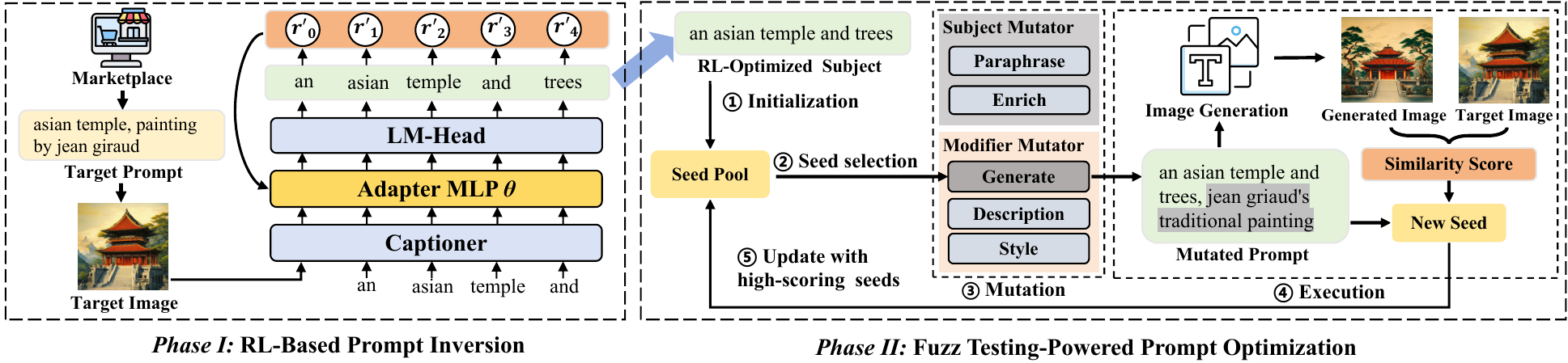} 
  \caption{Overview of \sys. Our method comprises two phases: (\textit{I}) a reinforcement learning–based optimization phase to reconstruct the primary subject, and (\textit{II}) a fuzzing-driven search phase to recover stylistic modifiers. }
  \label{fig:overview}
  \vspace{-1\baselineskip}
\end{figure*}
\section{Methodology}\label{sec:method}

\subsection{Threat Model}\label{sec:threat}

\noindent\textbf{Adversary's Goal.} Given a target image generated by a text-to-image generative model, the goal of prompt stealing is to invert the prompt used to generate the target image or a prompt that leads to highly similar images.
As such, the attack is typically evaluated by (1) image similarity between the generated image and the target image, including perceptual and semantic similarities;
(2) prompt similarity (or textual alignment) between the inverted and original prompts, measuring the semantic similarity at token- and sentence-levels.

\noindent\textbf{Adversary's Capability.} 
We assume that the adversary is only provided with a target image produced by a text-to-image generative model and has black-box access to the target text-to-image generative model. 
Additionally, the adversary does not have access to large-scale image-caption datasets or to prompt base with commercially traded prompts. 

\noindent\textbf{Defender's Capability.} 
The defenders are primarily focused on protecting high-quality prompts from being stolen. 
We assume they can apply post-hoc transformations to images generated with protected prompts before those images are disseminated. 

\subsection{Overview}\label{sec:overview}
We introduce an effective prompt stealing attack against text-to-image generative models called \sys, consisting of two main phases. 
In the \textit{phase I}, we formulate the prompt inversion as a Markov Decision Process~(MDP) and solve it using reinforcement learning with shaped reward. 
In the \textit{phase II}, we perform a fuzz testing-powered Prompt Optimization to iteratively refine the RL-derived prompt, exploring richer descriptions and stylistic modifiers while maximizing similarity between generated and target images.
The overview of \sys is shown in \autoref{fig:overview}.

\subsection{RL-Based Prompt Inversion}\label{sec:inversion}

Recalling the autoregressive prompt generation in image captioning models, at each step, the model observes the current partially constructed prompt and a representation of the target image, chooses the next token to append. 
We can formulate this process as an MDP \(\mathcal{M} = (\mathcal{S}, \mathcal{A}, \mathcal{P}, \mathcal{R}, \gamma)\) and solve it using reinforcement learning~\cite{10.5555/3312046}.
The \textbf{state} \(s_t \in \mathcal{S}\) is defined as \(s_t = \{p_{0:t-1}, x\} \), where \(p_{0:t-1}\) is the partial prompt and \(x\) is the target image. 
The \textbf{action} \(a_t \in \mathcal{A}\) selects the next token \(p_t\) from the vocabulary; 
and the \textbf{transition} \(\mathcal{P}(s_{t+1} \! \mid \! s_t,a_t)\) deterministically appends the selected token to form the new prefix. The \(\gamma \) is the \textbf{discount factor}.
The episode terminates when an end-of-sequence token is produced, yielding a complete prompt \(p_{0:T}\) and a synthesized image \(\hat{x}\), \(T\) is the sequence length determined when the model emits an end-of-sequence token.
The objective is to learn a policy \(\pi_\theta(a_t \! \mid \! s_t)\) that maximizes the expected discounted return \(\mathbb{E}_{\pi_\theta}\!\left[\sum_{t=0}^{T} \gamma^t R(s_t, a_t)\right]\).

\noindent \textbf{Reward Design.} Since our goal is to recover a prompt whose synthesized image \(\hat{x}\) is as close as possible to the target image \(x\), a naive design of \textbf{reward} \(\mathcal{R}\) is to measure the similarity between the generated image \(\hat{x}\) and the target image \(x\) directly. 
In specific, after generating the complete prompt \(p_{0:T}\), we synthesize an image \(\hat{x}\) with the target text-to-image generative model and compute the reward \(r_t = \mathcal{R}(s_t, a_t)\) as:
\begin{equation}
r_t =
\begin{cases}
    0, & 0 \le t < T, \\
    \Psi(x, \hat{x}), & t = T,
\end{cases}
\end{equation}

where \(\Psi(\cdot)\) is the cosine similarity between the generated image \(\hat{x}\) and the target image \(x\). It is calculated with an image encoder \(f_{\text{img}}(\cdot)\) from CLIP as:
\begin{equation}\label{eq:image-image-sim}
    \Psi(\hat{x}, x) \;=\;
    \frac{f_{\text{img}}(\hat{x}) \cdot f_{\text{img}}(x)}
         {\|f_{\text{img}}(\hat{x})\| \, \| f_{\text{img}}(x) \| }.
\end{equation} 

We invoke the target text-to-image generative model only after a complete prompt has been produced, so no environment feedback is available at intermediate steps, and we set \(r_t=0\) for \(0\le t < T\).

However, such reward \(r_t\) based only on image similarity leads to sparse supervision, as the agent receives feedback only at the end of each episode. 
This makes optimization inefficient and unstable, especially for long prompts with large action spaces.  
To address this, we utilize \textbf{potential-based reward shaping~\cite{ng1999policy}} to provide additional intermediate rewards \(r'_t\) for denser learning signals. 
The overall \textbf{shaped reward} \(r'_t = \mathcal{R'}(s_t,a_t)\) is defined as:
\begin{equation}
r'_t =
\begin{cases}
    \gamma\Phi(s_{t+1}) - \Phi(s_t), & 0 \le t < T, \\
    r_t - \Phi(s_t), & t = T,
\end{cases}
\end{equation}
where \(\Phi(\cdot)\) is the potential function. Specifically, we use CLIP text-image similarity to define it as:
\begin{equation}
    \Phi(s_t) \;=\; 
    \beta\cdot\frac{f_{\text{text}}(p_{1:t}) \cdot f_{\text{img}}(x)}
         {\|f_{\text{text}}(p_{1:t})\| \, \|f_{\text{img}}(x)\|},
\end{equation}
where \(\beta \) is a scaling coefficient, \(f_{\text{text}}(\cdot)\) and \(f_{\text{img}}(\cdot)\) are the CLIP text and image encoders that map prompts and images into a shared embedding space.

This reward design integrates dense potential-based shaping with structured terminal feedback, resulting in faster convergence and more stable optimization compared to sparse end-only rewards. 
Theoretically, as shown in \autoref{sec:reward_shaping}, the potential-based formulation preserves the optimal policy, ensuring that shaping affects only the learning dynamics rather than the underlying objective. 
Empirically, as shown in \autoref{fig:shape_vs_sparse}, the proposed potential-based reward shaping accelerates convergence and improves training efficiency while maintaining stability.

\noindent\textbf{Model Architecture.}
As shown in~\autoref{fig:overview}, we build our agent on top of a frozen
captioner. 
Specifically, when the captioner outputs a hidden state \(h_t \in \mathbb{R}^d\) at step \(t\), we introduce a trainable adapter \(\theta \) that parameterizes the policy:
\begin{equation}
    \tilde{h}_t = \theta(h_t),
\end{equation}
where \(\tilde{h}_t\) is then fed into the \emph{frozen} LM head of the captioner to obtain token logits and the action distribution. 

Following prior works that leverage imitation or supervised pretraining to stabilize and accelerate subsequent reinforcement learning \citep{Hester2018DQfD,Rajeswaran2018DAPG,Silver2016AlphaGo,Ouyang2022InstructGPT}, our training process consists of two key stages: \textit{(1)} imitation learning (IL) to warm-start the policy using expert trajectories generated by a captioner, and \textit{(2)} RL fine-tuning with reward shaping. 

\noindent\textbf{Imitation Learning Warm-start.}
We first perform imitation learning to initialize the adapter. 
Concretely, we sample expert token sequences \( \{(y_1,\dots,y_T)\} \) from the frozen captioner and replay them to collect hidden states \( \{h_t\} \). 
The adapter \(\theta \) is then trained with a standard cross-entropy objective against the next-token targets, using the \emph{frozen} LM head to produce logits:

\begin{equation}
    \mathcal{L}_{\text{IL}}
    = -\frac{1}{N}\sum_{(h_t, y_{t+1})}
    \log P_{\text{LM}}\!\big(y_{t+1}\;\big|\;\tilde{h}_t\big),
\end{equation}
where \(P_{\text{LM}}(\cdot \mid \tilde{h}_t)\) denotes the token distribution output by the frozen LM head given the adapted hidden state \(\tilde{h}_t\). 
Only the adapter parameters \(\theta \) are updated in this stage; the backbone and LM head remain frozen. 
This warm start provides a strong semantic initialization that stabilizes and accelerates subsequent RL training.

\noindent\textbf{PPO Optimization.}
We second train the adapter \(\theta \) with Proximal Policy Optimization (PPO)~\cite{schulman2017proximalpolicyoptimizationalgorithms}, using the clipped surrogate objective:
\begin{equation}
\mathcal{L}_{\text{PPO}} =
\mathbb{E}_t \Big[
\min\big(
\rho_t(\theta) A_t,\;
\text{clip}(\rho_t(\theta), 1 - \epsilon, 1 + \epsilon) A_t
\big)
\Big],
\end{equation}
where \(\rho_t(\theta)=\frac{\pi_\theta(a_t|s_t)}{\pi_{\theta_{\text{old}}}(a_t|s_t)}\) is the importance weight between the new and old policies, \(A_t\) is the advantage estimate, and \(\epsilon \) is the clipping parameter. 
\(\pi_\theta(a_t|s_t)\) denotes the probability of taking action \(a_t\) under state \(s_t\) with parameters \(\theta \). 
\(A_t\) is computed using the estimated state value from the value head, which guides the policy update toward actions that yield higher-than-expected rewards. 
The clip function truncates the policy ratio \(\rho_t(\theta)\) within the range \([1-\epsilon,\,1+\epsilon]\) 
to prevent excessively large policy updates, thereby stabilizing training and avoiding performance collapse.
Only the adapter and value head are updated during PPO optimization.

\subsection{Fuzz Testing-Powered Prompt Optimization}\label{sec:optimization}

The RL-optimized prompts successfully achieve high semantic alignment with the target image and precisely capture the core visual entities, yielding concise, content-faithful subjects. 
Nevertheless, their ability to reconstruct stylistic and compositional modifiers is limited, largely because image captioning backbones provide weak prior knowledge of modifiers training data. 
To address this limitation, we introduce a fuzz optimization to refine the RL-derived prompts. 
As shown in~\autoref{fig:overview} and Algorithm~\autoref{alg:fuzz_opt}, we initialize the seed pool with RL-optimized prompts and iteratively improve them through mutation and evaluation. 
At each iteration, a set of mutators, either subject- (e.g., paraphrasing, enrichment) or modifier-level (e.g., description and style generation), propose new prompt variants.
Each mutated prompt is then used to synthesize an image, whose similarity to the target image determines its score. 
High-scoring prompts are retained in the seed pool for further mutation, enabling exploration and exploitation of the prompt space in a self-improving loop. 
This process progressively enriches the prompt with expressive modifiers while preserving its semantic grounding.
However, applying fuzz testing to prompt stealing faces two key challenges.
\textbf{First, there is a lack of high-quality seeds.} 
Recent studies~\cite{gong2025papillon, 10.1145/3460319.3464795, 9787966} show that seed choice profoundly shapes the efficacy and trajectory of fuzzing, and sparse or weak seeds lead to poor coverage and premature convergence. 
\textbf{Second, effective mutation strategies are missing.} 
Existing approaches, such as traditional mutation strategies like Havoc in AFL~\cite{zalewski2014afl} or recent LLM-based text mutation methods~\cite{gong2025papillon, yu2024gptfuzzerredteaminglarge}, are not tailored to structured text-to-image prompts with \emph{subject} and \emph{modifier}. 

To address these challenges, we introduce two key designs that make fuzzing both structure-aware and sample-efficient. 
First, we guide exploration with a compact, quality-driven seed pool that retains strong candidates and schedules the next mutation via principled selection.
Second, we mutate prompts with strategies that respect the compositional structure of text-to-image inputs, separately handling \emph{subject} and \emph{modifier} to enrich details without drifting from content.

\noindent\textbf{Seed Update and Selection.}
\sys maintains a \textit{capacity-constrained elitist} seed pool with a fixed size of candidates. 
After each evaluation, the new prompt and its score are inserted into the pool; the pool is kept sorted by score, and the lowest-scoring prompt is evicted when the capacity is exceeded. 
To choose the next seed for mutation, we adopt the Monte Carlo Tree Search (MCTS)~\cite{10.1007/978-3-540-75538-8_7} algorithm to balance exploration and exploitation. This mechanism preserves high-quality prompts, continuously explores promising neighborhoods in prompt space.

\noindent\textbf{Hybrid-Strategy Prompt Mutation.} To enhance expressiveness and visual fidelity, \sys employs a VLM mutator (e.g., Qwen2-VL-2B-Instruct~\cite{Qwen2VL}) with five targeted operators: (1) \textbf{\emph{Subject-Paraphrase}} rewrites the subject for more natural phrasing without changing meaning; 
(2) \textbf{\emph{Subject-Enrich}} inserts brief image-grounded details (e.g., color, count, pose) while preserving sentence structure; 
(3) \textbf{\emph{Modifier-Generate}} jointly produces a new description and style from the image and subject; 
(4) \textbf{\emph{Modifier-Description}} refines the existing description with spatial relations, composition, and lighting cues; 
and (5) \textbf{\emph{Modifier-Style}} adjusts medium, texture, lens, and quality tokens to strengthen aesthetic control. 
Together, these operators expand RL-derived subjects into well-formed prompts that remain semantically grounded yet stylistically rich.
See~\autoref{sec:vlm_mutator_prompts} for detailed descriptions and the exact prompt templates of each mutator.
\begin{algorithm}[t]
\caption{Workflow of Fuzz Optimization}\label{alg:fuzz_opt}

\textbf{Input:} Target image $\bm x$; RL-optimized prompt $p_{\mathrm{RL}}$; 
text-to-image generator $\mathcal{G}$; similarity function $\Psi$; 
VLM-based mutator set $\mathcal{M}$; 
seed pool capacity $K$; query budget $Q$. \\
\textbf{Result:} Optimized subject--modifier prompt $p^\star$.

\begin{algorithmic}[1]

\State $\mathcal{S} \gets \{(p_{\mathrm{RL}},\, \Psi(\mathcal{G}(p_{\mathrm{RL}}), \bm x))\}$ \Comment{\underline{\textit{Initialization}}}

\For{$t = 1$ \textbf{to} $Q$}
    \State $(p_{\mathrm{seed}}, s_{\mathrm{seed}}) \gets \textsc{SelectFromPool}(\mathcal{S})$ \Comment{\underline{\textit{MCTS}}}
    \State $\mathcal{M}_i \gets \textsc{SampleMutator}(\mathcal{M})$ 
    \Comment{\underline{\textit{Mutation}}}
    \State $p_{\mathrm{new}} \gets \mathcal{M}_i(p_{\mathrm{seed}}, \bm x)$ 
    \State $\hat{\bm x} \gets \mathcal{G}(p_{\mathrm{new}})$ \Comment{\underline{\textit{Execution}}}
    \State $s_{\mathrm{new}} \gets \Psi(\hat{\bm x}, \bm x)$
    \State $\mathcal{S} \gets \mathcal{S} \cup \{(p_{\mathrm{new}}, s_{\mathrm{new}})\}$ \Comment{\underline{\textit{Update}}}
    \State keep top-$K$ elements in $\mathcal{S}$ by score $s$
\EndFor

\State $(p^\star, s^\star) \gets \arg\max_{(p,s)\in\mathcal{S}} s$
\State \textbf{return} $p^\star$
\end{algorithmic}
\end{algorithm}

\begin{table*}[t]
\centering
\caption{Image similarity and textual alignment comparison across datasets.}
\resizebox{\linewidth}{!}{
\begin{tabular}{ll*{12}{c}}
\toprule
\multirow{2}{*}{\textbf{Dataset}} & \multirow{2}{*}{\textbf{Method}} & 
\multicolumn{3}{c}{\textbf{Stable Diffusion v1.5}} & 
\multicolumn{3}{c}{\textbf{SDXL-Turbo}} & 
\multicolumn{3}{c}{\textbf{FLUX.1 dev}} & 
\multicolumn{3}{c}{\textbf{Stable Diffusion 3.5 Medium}} \\
\cmidrule(lr){3-5} \cmidrule(lr){6-8} \cmidrule(lr){9-11} \cmidrule(lr){12-14}
 & & CLIP$\uparrow$ & LPIPS$\downarrow$ & SBERT$\uparrow$ & CLIP$\uparrow$ & LPIPS$\downarrow$ & SBERT$\uparrow$ & CLIP$\uparrow$ & LPIPS$\downarrow$ & SBERT$\uparrow$ & CLIP$\uparrow$ & LPIPS$\downarrow$ & SBERT$\uparrow$ \\
\midrule
\multirow{7}{*}{MS COCO} 
 & BLIP            &0.858&0.458&\underline{0.660} &0.855&0.444&0.643 &0.885&0.467&\textbf{0.700} &0.881&0.409&\underline{0.718} \\
 & CLIP-IG         &\underline{0.874}&0.461&0.499 &0.862&0.466&0.486 &0.887&0.491&0.551 &0.883&0.461&0.565 \\
 & VLM-as-expert      &0.861&0.466&0.556 &\underline{0.866}&0.462&0.539 &\underline{0.923}&\underline{0.407}&0.588 &\underline{0.913}&\underline{0.387}&0.612 \\
 & PH2P            &0.673&0.724&0.041 &0.645&0.612&0.047 &0.641&0.703&0.040 &0.642&0.724&0.044 \\
 & PromptStealer   &0.861&\underline{0.453}&0.633 &0.861&\underline{0.439}&\underline{0.652} &0.898&0.464&0.680 &0.866&0.446&0.693 \\
 & VGD             &0.816&0.700&0.424 &0.802&0.565&0.410 &0.791&0.630&0.417 &0.785&0.691&0.416 \\
 & \cellcolor{blue!10}\sys~(Ours) 
   & \cellcolor{blue!10}\textbf{0.933} & \cellcolor{blue!10}\textbf{0.342} & \cellcolor{blue!10}\textbf{0.664}
   & \cellcolor{blue!10}\textbf{0.934} & \cellcolor{blue!10}\textbf{0.340} & \cellcolor{blue!10}\textbf{0.673}
   & \cellcolor{blue!10}\textbf{0.958} & \cellcolor{blue!10}\textbf{0.345} & \cellcolor{blue!10}\underline{0.683}
   & \cellcolor{blue!10}\textbf{0.953} & \cellcolor{blue!10}\textbf{0.303} & \cellcolor{blue!10}\textbf{0.751} \\
\midrule
\multirow{7}{*}{Flickr} 
 & BLIP            &0.773&0.493&\underline{0.547} &0.808&0.485&0.549 &0.833&0.499&\underline{0.620} &0.820&0.465&\underline{0.637} \\
 & CLIP-IG         &\underline{0.833}&\underline{0.466}&0.472 &\underline{0.835}&0.478&0.468 &0.860&0.508&0.511 &0.857&0.487&0.547 \\
 & VLM-as-expert      &0.817&0.469&0.514 &0.827&\underline{0.473}&0.501 &\underline{0.895}&\underline{0.447}&0.571 &\underline{0.874}&\underline{0.431}&0.578 \\
 & PH2P            &0.638&0.715&0.039 &0.616&0.629&0.046 &0.630&0.686&0.038 &0.613&0.721&0.030 \\
 & PromptStealer   &0.783&0.504&0.501 &0.816&0.475&\underline{0.581} &0.843&0.498&0.595 &0.835&0.484&0.622 \\
 & VGD             &0.764&0.671&0.341 &0.761&0.577&0.366 &0.764&0.613&0.353 &0.753&0.684&0.329 \\
 & \cellcolor{blue!10}\sys~(Ours)
   & \cellcolor{blue!10}\textbf{0.894} & \cellcolor{blue!10}\textbf{0.402} & \cellcolor{blue!10}\textbf{0.568}
   & \cellcolor{blue!10}\textbf{0.910} & \cellcolor{blue!10}\textbf{0.405} & \cellcolor{blue!10}\textbf{0.603}
   & \cellcolor{blue!10}\textbf{0.927} & \cellcolor{blue!10}\textbf{0.411} & \cellcolor{blue!10}\textbf{0.628}
   & \cellcolor{blue!10}\textbf{0.916} & \cellcolor{blue!10}\textbf{0.410} & \cellcolor{blue!10}\textbf{0.660} \\
\midrule
\multirow{7}{*}{Lexica} 
 & BLIP            &0.699&0.590&0.319 &0.749&0.507&0.397 &0.760&0.554&0.335 &0.746&0.560&0.406 \\
 & CLIP-IG         &\underline{0.840}&\underline{0.471}&\underline{0.553} &0.856&0.451&0.591 &0.860&0.504&0.587 &\underline{0.859}&0.474&0.616 \\
 & VLM-as-expert      &0.794&0.501&0.507 &0.811&\underline{0.450}&0.527 &\underline{0.866}&\underline{0.445}&0.535 &0.857&\underline{0.441}&0.572 \\
 & PH2P            &0.640&0.729&0.178 &0.662&0.569&0.171 &0.657&0.657&0.173 &0.613&0.682&0.174 \\
 & PromptStealer   &0.762&0.560&\textbf{0.563} &0.787&0.475&\textbf{0.594} &0.788&0.563&\textbf{0.600} &0.804&0.520&\underline{0.622} \\
 & VGD             &0.753&0.701&0.466 &0.766&0.530&0.446 &0.760&0.612&0.481 &0.762&0.643&0.485 \\
 & \cellcolor{blue!10}\sys~(Ours) 
   & \cellcolor{blue!10}\textbf{0.875} & \cellcolor{blue!10}\textbf{0.450} & \cellcolor{blue!10}\textbf{0.563}
   & \cellcolor{blue!10}\textbf{0.911} & \cellcolor{blue!10}\textbf{0.420} & \cellcolor{blue!10}\underline{0.591}
   & \cellcolor{blue!10}\textbf{0.921} & \cellcolor{blue!10}\textbf{0.435} & \cellcolor{blue!10}\underline{0.594}
   & \cellcolor{blue!10}\textbf{0.921} & \cellcolor{blue!10}\textbf{0.407} & \cellcolor{blue!10}\textbf{0.629} \\
\bottomrule
\end{tabular}}
\label{tab: image_sim_baselines}
\end{table*}

\begin{figure*}[t] 
  \centering
  \includegraphics[width=1.0\linewidth]{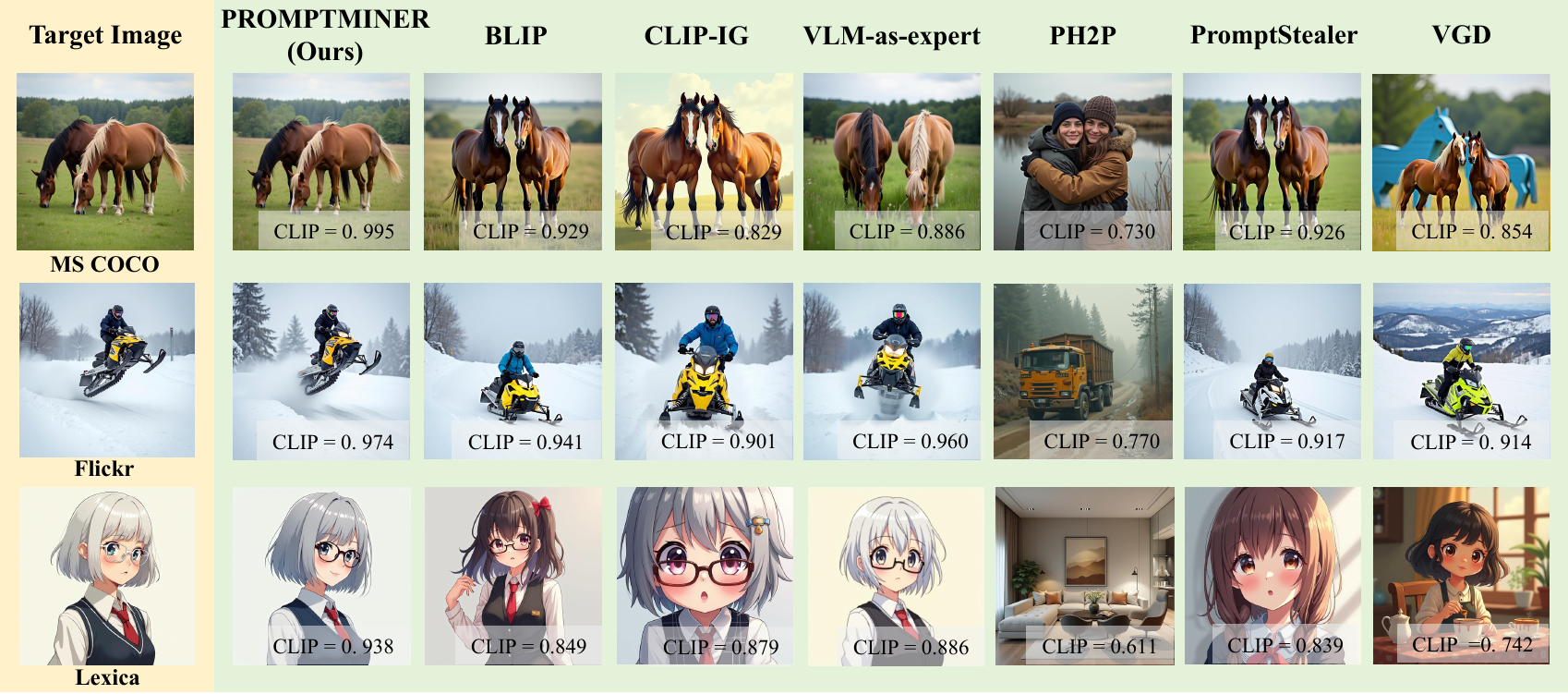} 
  \caption{Visualization of generated images compared with target image.}
  \label{fig:img_sim_visual}
    \vspace{-1\baselineskip}
\end{figure*}

\section{Evaluation}\label{sec:eval}
\subsection{Experiment Setup}
\noindent\textbf{Target Models.} We employ the most widely used state-of-the-art models, including Stable Diffusion v1.5~\cite{Rombach_2022_CVPR}, SDXL-Turbo~\cite{sauer2023adversarialdiffusiondistillation}, Stable Diffusion 3.5 Medium~\cite{esser2024scalingrectifiedflowtransformers} and FLUX.1 dev~\cite{flux2024}, as our target text-to-image generative models.

\noindent\textbf{Datasets.} 
We empirically evaluate our prompt stealing pipeline on four widely used datasets: MS COCO~\cite{10.1007/978-3-319-10602-1_48}, Flickr~\cite{young2014image}, Lexica~\cite{SQBZ24}, and DiffusionDB~\cite{wangDiffusionDBLargescalePrompt2022}. MS COCO and Flickr feature sentence-style captions that describe entities, actions, and scene layout without art-style tags, whereas Lexica and DiffusionDB contain keyword-driven prompts that follow a \textit{“subject, modifiers”} pattern with comma-separated stylistic and technical tokens. 
For the first three datasets, we randomly select 50 prompts and use the target text-to-image generators to synthesize the corresponding images. 
For DiffusionDB, we treat it as an \emph{in-the-wild} dataset: instead of generating new images with the target models, we directly use 50 original images from the dataset, where the underlying generative models are unknown.

\noindent\textbf{Baselines.} We compare \sys with state-of-the-art prompt stealing and auxiliary baselines: BLIP~\cite{li2022blip}, an image–captioning model producing sentence-style prompts; CLIP Interrogator~(CLIP-IG)~\cite{clip_interrogator_github}, which pairs CLIP with BLIP and selects modifiers from a large pool; VLM-as-expert using GPT-4o~\cite{hurst2024gpt} for multimodal prompt generation; PromptStealer~\cite{shen2024promptstealingattackstexttoimage}, which trains a subject generator and a modifier classifier on large labeled data; PH2P~\cite{ph2p2024cvpr}, which applies delayed discrete projection from late diffusion steps; and VGD~\cite{kim2025visuallyguideddecodinggradientfree}, a gradient-free method that uses an LLM with CLIP-based guidance to produce coherent prompts.
\begin{figure*}[t]
  \centering
  \begin{subfigure}[t]{0.48\linewidth}
    \centering
    \begin{subfigure}[t]{0.48\linewidth}
      \includegraphics[width=\linewidth]{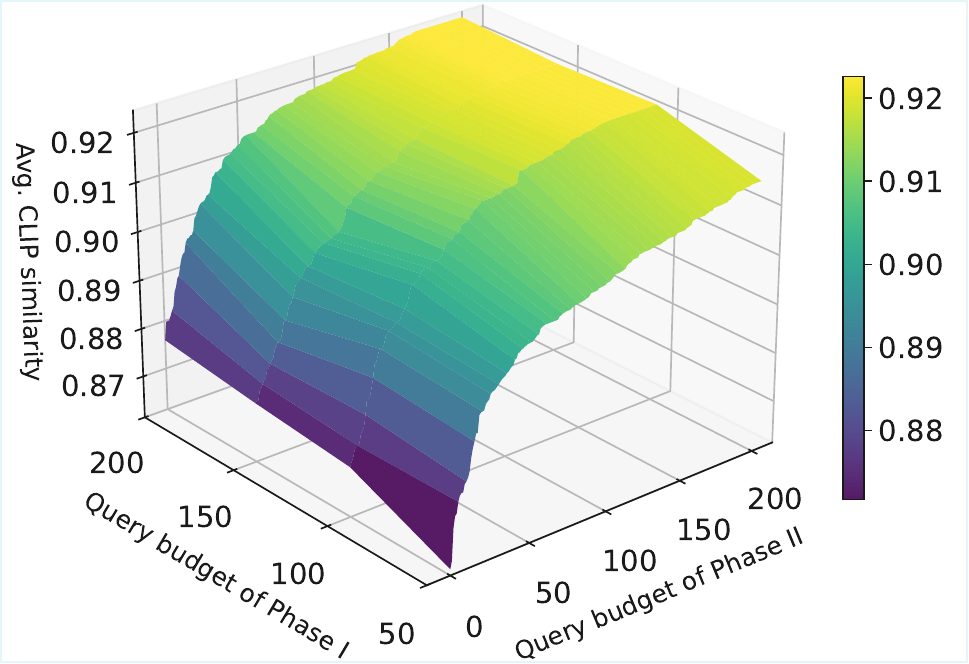}
    \end{subfigure}\hfill
    \begin{subfigure}[t]{0.48\linewidth}
      \includegraphics[width=\linewidth]{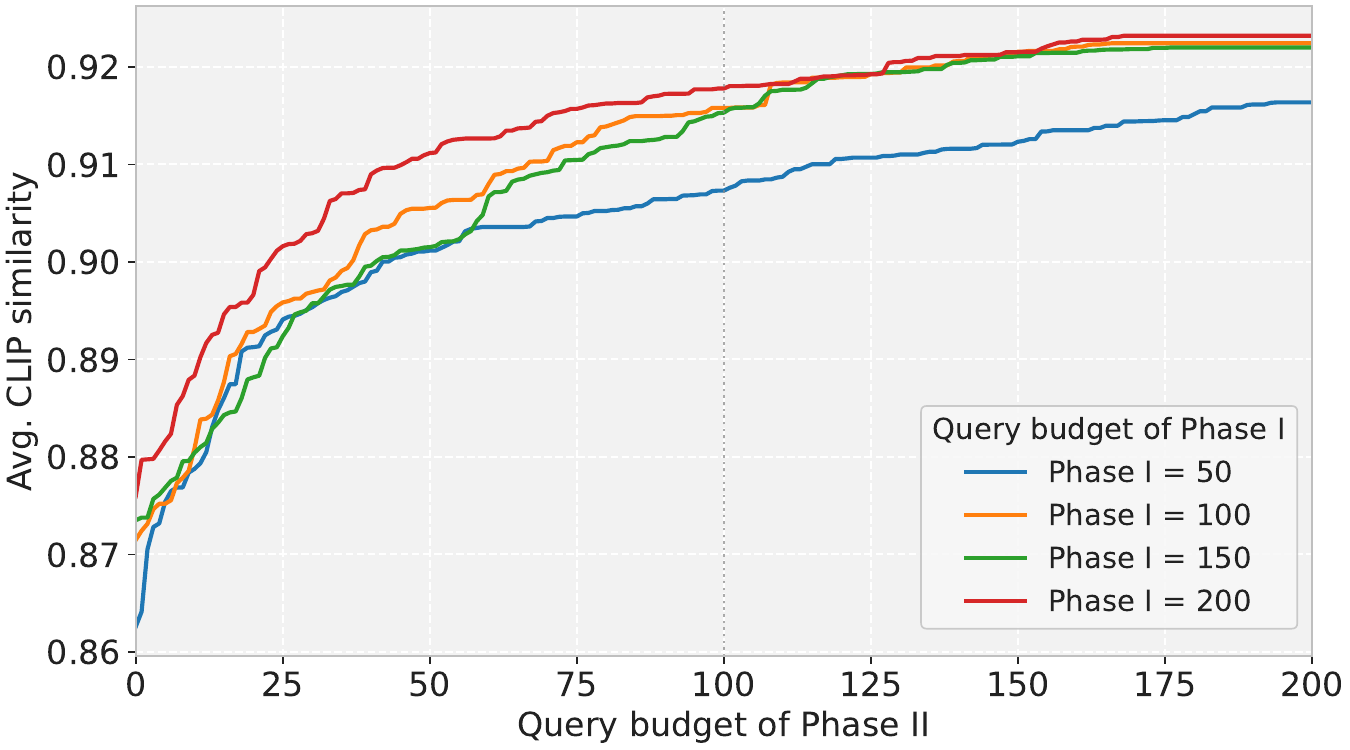}
    \end{subfigure}
    \caption{Impact of query budget on Flick dataset.}
    \label{fig:flickr_joint}
  \end{subfigure}\hfill
  \begin{subfigure}[t]{0.48\linewidth}
    \centering
    \begin{subfigure}[t]{0.48\linewidth}
      \includegraphics[width=\linewidth]{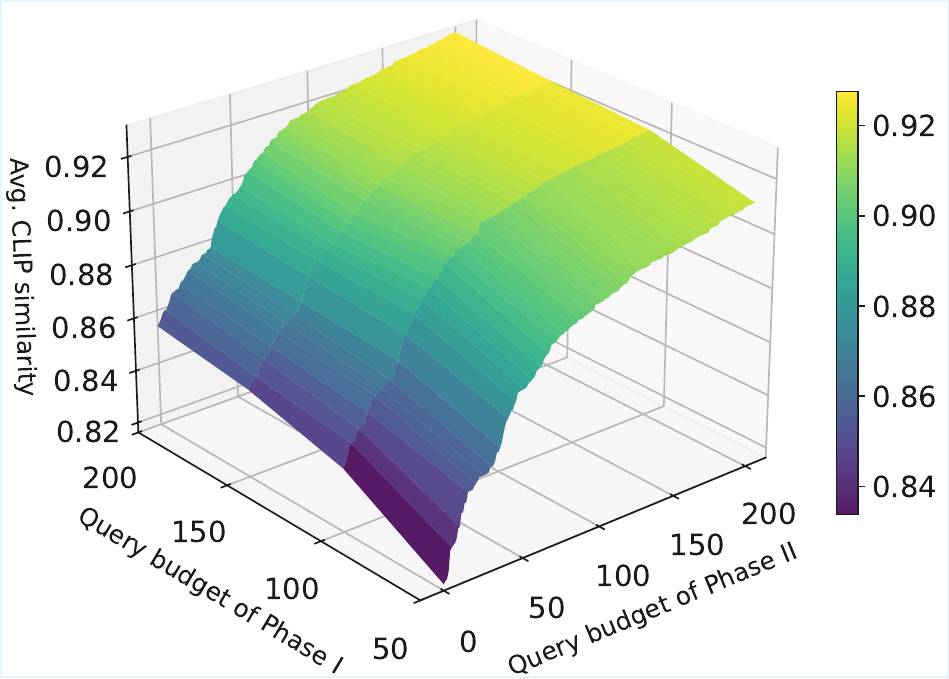}
    \end{subfigure}\hfill
    \begin{subfigure}[t]{0.48\linewidth}
      \includegraphics[width=\linewidth]{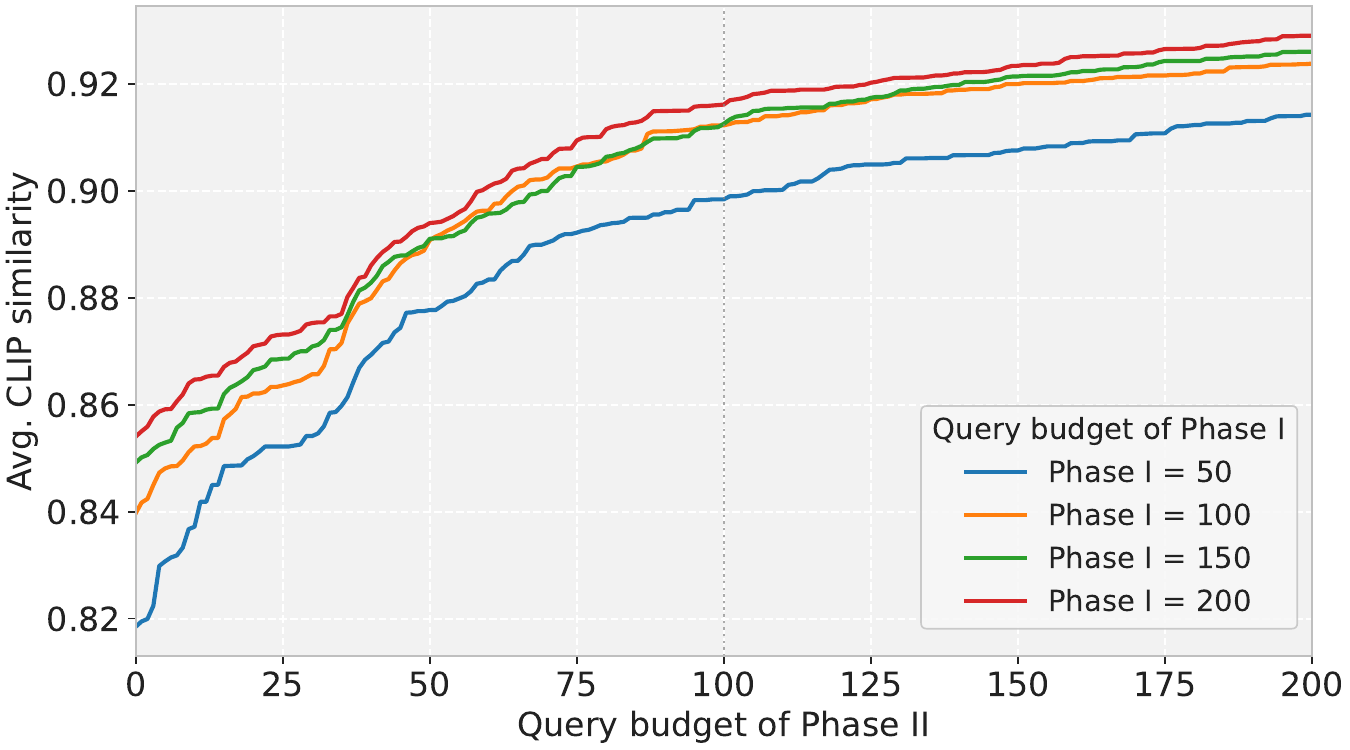}
    \end{subfigure}
    \caption{Impact of query budget on Lexica dataset.}
    \label{fig:lexica_joint}
  \end{subfigure}

  \caption{Impact of query budget on \textit{phase I} (RL-based Prompt Inversion) and \textit{phase II} (Fuzz Testing–Powered Prompt Optimization).}
  \label{fig:budget_fourpanel}

\end{figure*}

\noindent\textbf{Evaluation Metrics.} We evaluate prompt stealing performance in terms of \textbf{image similarity} and \textbf{textual alignment}. For image similarity, we use CLIP~\cite{pmlr-v139-radford21a} and LPIPS~\cite{Zhang_2018_CVPR} similarity, where CLIP similarity measures high-level semantic alignment in a joint vision–language space and LPIPS similarity captures low-level perceptual differences. For textual alignment, we use Sentence-BERT (SBERT)~\cite{reimers2019sentencebertsentenceembeddingsusing} to compute the cosine similarity between embeddings of the inverted and ground-truth prompts, indicating how well the recovered text preserves the original semantics. 

More details about the experimental setup are provided in the supplementary material. 

\subsection{Experiment Result}

\begin{table}[t]
\centering
\caption{Prompt stealing performance on \textit{in-the-wild} images.}
\scriptsize
\begin{tabular}{lcccc}
\toprule
\multirow{2}{*}{\textbf{Method}} & 
\multicolumn{2}{c}{\textbf{Image Similarity}} & 
\multicolumn{1}{c}{\textbf{Textual Alignment}} \\
\cmidrule(lr){2-3} \cmidrule(lr){4-4}
 & CLIP$\uparrow$ & LPIPS$\downarrow$ & SBERT$\uparrow$ \\
\midrule
BLIP            &0.753  &0.659  &0.365  \\
CLIP-IG         &\underline{0.803}  &0.628  &0.541  \\
VLM-as-expert      &0.782  &\underline{0.627}  &\underline{0.544}  \\
PH2P            &0.648  &0.706  &0.154  \\
PromptStealer   &0.754  &0.643  &0.542  \\
VGD             &0.749  &0.649  &0.454  \\
\cellcolor{blue!10}\sys~(Ours) 
   &\cellcolor{blue!10}\textbf{0.863} & \cellcolor{blue!10}\textbf{0.622}
   & \cellcolor{blue!10}\textbf{0.545}\\
\bottomrule
\end{tabular}
\label{tab:inwild_image_sbert}
\vspace{1\baselineskip}
\end{table}

\noindent \textbf{Image Similarity.} We evaluate the effectiveness of the inverted prompts generated by \sys for image generation. As shown in~\autoref{tab: image_sim_baselines}, \sys surpasses all the baselines across all datasets and text-to-image generative models in both CLIP and LPIPS similarity. Concretely, on MS~COCO, \sys attains CLIP similarity (0.958) and LPIPS similarity (0.345) with FLUX.1 dev, and CLIP Similarity (0.953) and LPIPS Similarity (0.303) with Stable Diffusion 3.5 Medium; on Flickr with FLUX.1 dev, \sys records CLIP similarity (0.927) together with LPIPS similarity (0.411); on Lexica with Stable Diffusion 3.5 Medium, it achieves CLIP similarity (0.921) and LPIPS similarity (0.407). 
These gains demonstrate state-of-the-art image similarity, reflecting stronger semantic alignment and perceptual fidelity than prior methods. 
We visualize representative generations on FLUX.1 dev across MS COCO, Flickr, and Lexica in ~\autoref{fig:img_sim_visual}. The images produced by \sys are visibly closer to the targets than those from baselines, consistent with the quantitative gains. Additional examples are provided in the ~\autoref{sec:qualitative_results}.

\noindent \textbf{Textual Alignment.} We measure the semantic alignment between the inverted prompts and the ground-truth prompts using sentence-BERT Socre~(SBERT)~\cite{reimers2019sentencebertsentenceembeddingsusing}. \sys achieves the highest SBERT across nearly all datasets and models, showing superior text-level recovery. With Stable Diffusion 3.5 Medium,  SBERT rises from 0.718 with BLIP on MS~COCO to \textbf{0.751} with \sys; on Flickr, it improves from 0.637 to \textbf{0.660}; on Lexica, \sys reaches \textbf{0.629}, surpassing all baselines. These consistent improvements confirm that \sys reconstructs subject–modifier semantics more faithfully, producing linguistically natural and semantically aligned prompts.

We attribute these improvements to the two-stage design of \sys. The RL-based adapter ensures that the recovered subjects remain tightly aligned with image semantics, providing a stable content foundation. The subsequent fuzz testing-powered optimization stage systematically explores modifier space, enriching the prompt with stylistic and compositional cues that are crucial for achieving perceptual resemblance. This synergy allows \sys to generate inverted prompts that not only reproduce the main subject accurately but also replicate fine-grained artistic attributes.

\noindent \textbf{Prompt Stealing on In-The-Wild Images.} 
We further evaluate the performance of \sys on real-world images whose underlying generation models are unknown, demonstrating the generalization ability of our method in unconstrained scenarios. 
Specifically, we randomly sample 50 image–prompt pairs from the DiffusionDB~\cite{wangDiffusionDBLargescalePrompt2022} dataset, where each image is generated by an unknown diffusion model and accompanied by its original text prompt. As shown in~\autoref{tab:inwild_image_sbert}, our method achieves the highest scores on all metrics.
This demonstrates that even when the source model of the image is completely unknown, our method can effectively leverage a proxy text-to-image generative model to perform reliable prompt stealing. 
Such generalization highlights the robustness of our framework and its ability to capture transferable visual–textual correlations across heterogeneous generation models.

\begin{figure*}[t]
  \centering
  \makebox[\textwidth][c]{%
   \begin{minipage}[t]{0.42\textwidth}
      \vspace{0pt}\centering
      \captionof{table}{Potential defenses against \sys.}
      \label{tab:defense_results}
      \scriptsize
      \begin{tabular}{lccc}
        \toprule
        \multirow{2}{*}{\textbf{Defense}} &
        \multicolumn{2}{c}{\textbf{Image Similarity}} &
        \multicolumn{1}{c}{\textbf{Textual Alignment}} \\
        \cmidrule(lr){2-3}\cmidrule(lr){4-4}
         & CLIP$\uparrow$ & LPIPS$\downarrow$ & SBERT$\uparrow$ \\
        \midrule
        No defense             & 0.911 & 0.420 & 0.591 \\
        Random Noise Injection & 0.903 & 0.420 & 0.572 \\
        Puzzle Effect          & 0.902 & 0.425 & 0.561 \\
        Textual Watermarking   & 0.887 & 0.445 & 0.583 \\
        \bottomrule
      \end{tabular}
    \end{minipage}%
    \hspace{0.02\textwidth}%
    \begin{minipage}[t]{0.27\textwidth}
      \vspace{0pt}\centering
      \includegraphics[width=\linewidth]{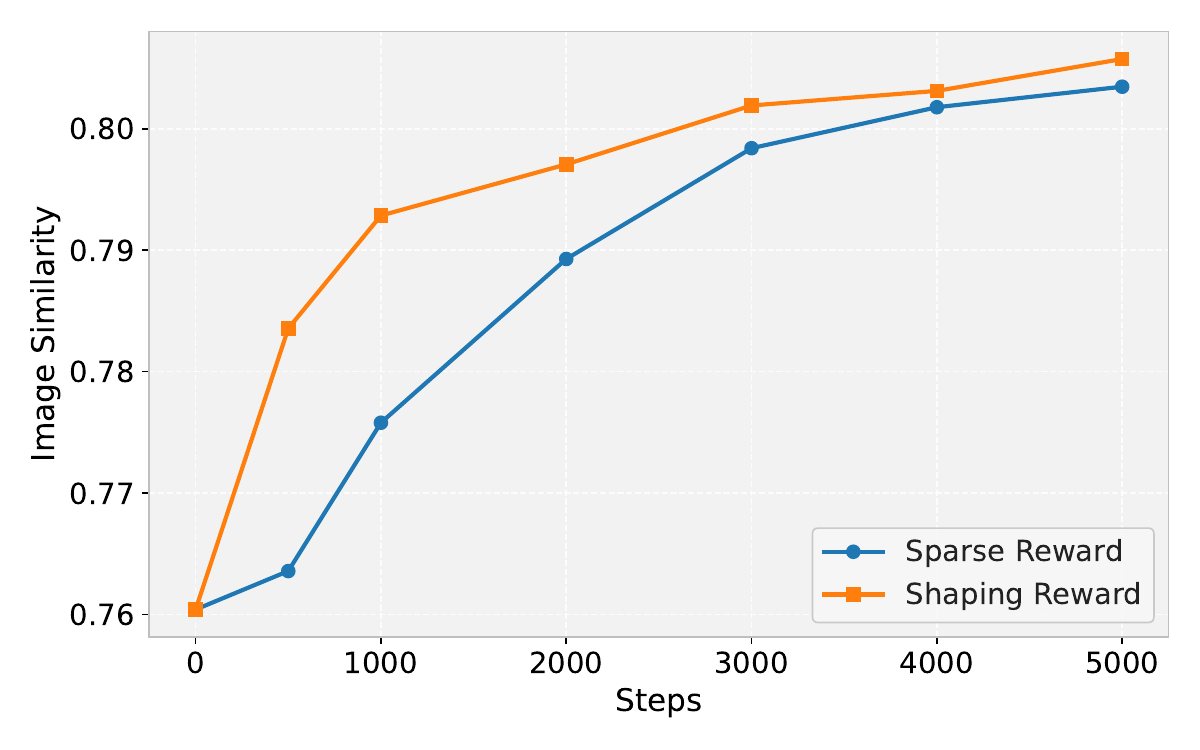}
      \captionof{figure}{Impact of reward shaping. 
      }
      \label{fig:shape_vs_sparse}
    \end{minipage}%
    \hspace{0.01\textwidth}
    \begin{minipage}[t]{0.30\textwidth}
      \vspace{0pt}\centering
      \includegraphics[width=\linewidth]{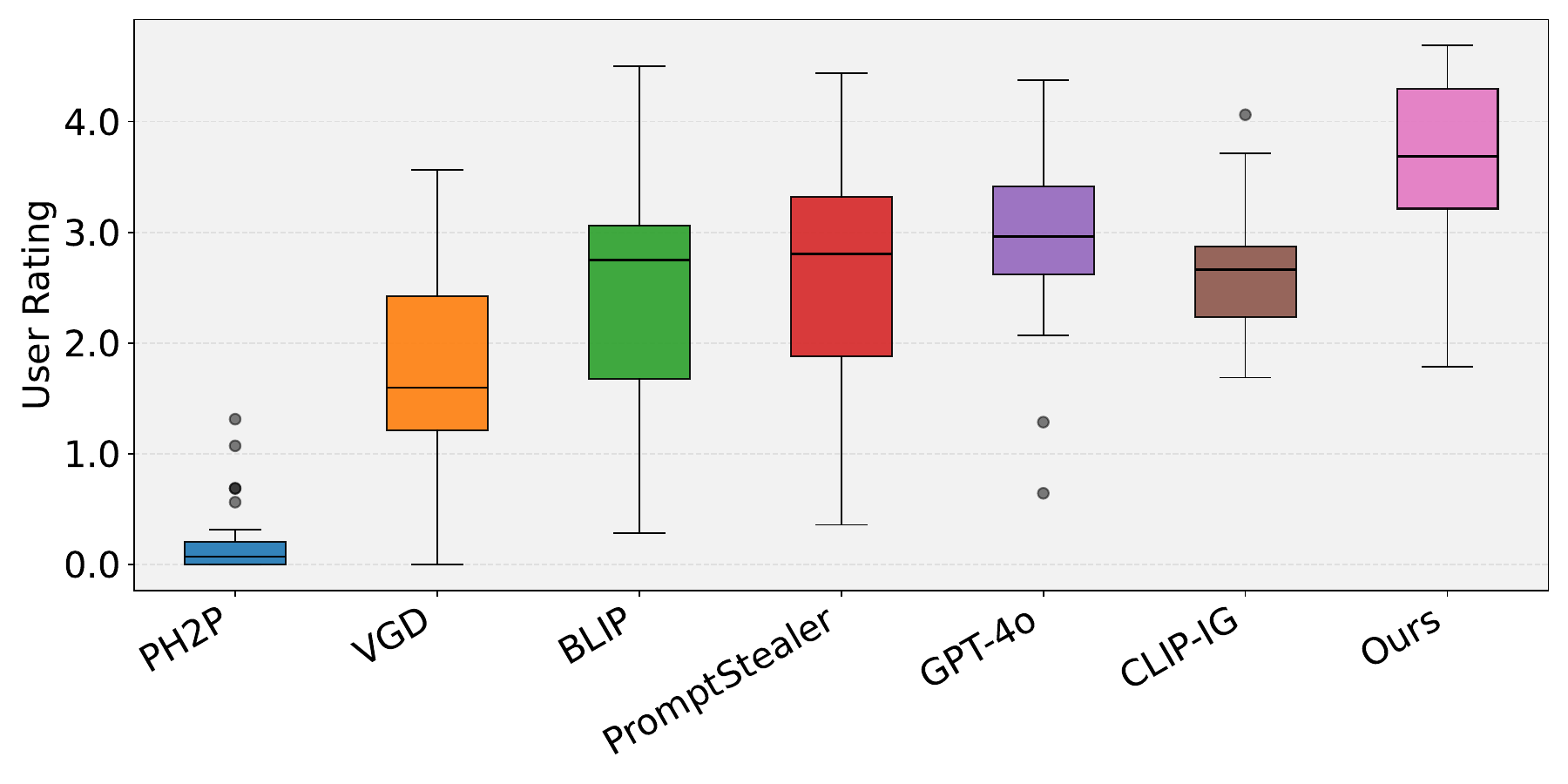}
      \captionof{figure}{Ratings from the user study.}
      \label{fig:user_study}
    \end{minipage}%


  }
\end{figure*}

\subsection{Ablation Study}

\noindent \textbf{Impact of Query Budget.}
\autoref{fig:budget_fourpanel} illustrates the performance of \sys on CLIP similarity under different query budgets. Specifically, \emph{Phase~I} (RL-based Prompt Inversion) uses query budgets of 50, 100, 150, and 200, while \emph{Phase~II} (Fuzz Testing–Powered Prompt Optimization) varies its query budget from 1 to 200. 
On Flickr (predominantly subject-only prompts), \emph{Phase I} exhibits rapid convergence: once the query budget reaches roughly one hundred, additional budget yields minimal improvement. \emph{Phase II} attains comparable final CLIP scores across these settings—indicating limited marginal value from further \emph{Phase I} investment for simple prompts.
On Lexica (prompts combining subjects with rich modifiers), the enlarged search space slows \emph{Phase I} convergence; allocating a larger budget produces stronger initial subjects and enables \emph{Phase II} to reach higher final CLIP similarity.
Considering both efficiency and quality, \textbf{there exists a clear trade-off between allocating queries to \emph{Phase~I} and \emph{Phase~II}}. As shown in~\autoref{fig:budget_fourpanel}, the optimal goal is to reach the \textbf{upper yellow region}—representing high CLIP similarity—with minimal query expenditure. Increasing the \emph{Phase~I} budget enhances initialization quality, particularly for complex prompts, whereas investing more in \emph{Phase~II} refines stylistic modifiers but quickly exhibits diminishing returns. Empirically, balanced allocations with \textbf{100} queries per phase achieve strong performance efficiently, approaching the optimal region without excessive cost.

\noindent \textbf{Impact of Reward Shaping.} We evaluate the impact of reward shaping by comparing training dynamics under sparse versus shaping rewards. The sparse reward provides feedback only at the end of an episode, while the shaped reward adds intermediate dense signals to accelerate learning. As shown in \autoref{fig:shape_vs_sparse}, 
The shaped reward supplies dense intermediate feedback via a potential function, enabling faster progress in early stages and reaching a high image similarity with substantially fewer steps.
In contrast, the sparse-reward variant lacks intermediate guidance and therefore converges more slowly.
Overall, potential-based reward shaping accelerates convergence and improves training efficiency while maintaining stability. 

For additional ablation results, please see~\autoref{sec:app_ablation}.

\subsection{User Study}
We conducted a user study to further evaluate the effectiveness of our method from the perspective of human perception. Participants are presented with a target image and several anonymous candidate images generated by \sys and baselines, and are asked to select the one that appears most visually similar to the target. The similarity preference is rated on a five-point scale from 0 to 5. We randomly select a total of 30 test cases from the experimental results covering all the three datasets and four generative models. According to the results in~\autoref{fig:user_study}, \sys consistently receives the highest preference scores, indicating that its recovered prompts generate images that better align with human visual perception.

\section{Potential Defenses}\label{sec:defences}
Given the increasing risk of prompt stealing attacks, we further explore several existing defenses that can be applied during image dissemination to mitigate potential prompt extraction or replication. 
Specifically, as same as~\cite{zhao2025effectivepromptstealingattack}, we investigate three representative defenses: (1) \textbf{Random noise Injection}, where Gaussian noise 
is added to subtly perturb pixel-level information; (2) \textbf{Puzzle effect}, which divides the image into a $4\times4$ grid and applies random local translations 
which is applied in practice~\cite{Dianamoran@PromptBase}; and (3) \textbf{Textual watermarking}, which overlays a visible pattern such as “@watermark” across the image 
to indicate ownership or provenance and also used in reality~\cite{Mylab@PromptBase}. These methods differ in their balance between imperceptibility and robustness: while noise and puzzle effect are visually subtle and suitable for general protection, watermarking provides explicit attribution at the cost of stronger visual alteration. We evaluate the effectiveness of each defense in hindering prompt stealing in \hyperref[tab:defense_results]{Table~\ref*{tab:defense_results}}. It is obvious that these post-processing defenses have only a minor impact on preventing our prompt stealing attack, indicating that \sys is inherently resilient to low-level visual perturbations due to its semantics-driven optimization process.

\section{Conclusion}\label{sec:conclusion}
In this paper, we propose \sys, a novel and effective black-box framework for prompt stealing against text-to-image generative models. Unlike prior approaches that rely on gradient access or large-scale supervised training, \sys efficiently reconstructs prompts through a two-phase process, RL–based subject inversion and fuzz-testing–powered modifier optimization, achieving both semantic precision and stylistic fidelity in the recovered prompts. Extensive experiments across diverse datasets and diffusion backbones show that \sys consistently outperforms existing baselines in image and textual similarity, generalizes to in-the-wild settings, and remains robust under common defensive perturbations. This work highlights the growing need for prompt protection in generative systems and provides a foundation for future research on secure and interpretable prompt recovery.
{
	\small
	\bibliographystyle{ieeenat_fullname}
	\bibliography{main}
}
\clearpage
\newpage

\suppstart{Appendix}

\section{More Details of the Used Models}
\label{sec:models}
In this section, we provide more details about the models used in the experiments. We conduct experiments on four representative text-to-image generative models with different architectural designs, including Stable Diffusion v1.5\citep{Rombach_2022_CVPR}, SDXL Turbo~\cite{sauer2023adversarialdiffusiondistillation}, Stable Diffusion 3.5 Medium~\cite{esser2024scalingrectifiedflowtransformers}, and FLUX.1 dev~\cite{flux2024}.

\begin{itemize}
\item \noindent\textbf{Stable Diffusion v1.5:} This model is initialized with the weights of the Stable-Diffusion-v1.2 checkpoint and subsequently fine-tuned on 595k steps at resolution 512x512 on "laion-aesthetics v2.5+" and 10\% dropping of the text-conditioning to improve classifier-free guidance sampling. The text encoder of this model is CLIP-ViT/L.

\item \noindent\textbf{SDXL Turbo:} This model is a distilled version of SDXL 1.0, trained for real-time synthesis. SDXL-Turbo is based on a novel training method called Adversarial Diffusion Distillation (ADD), which allows sampling large-scale foundational image diffusion models in 1 to 4 steps at high image quality. This approach uses score distillation to leverage large-scale off-the-shelf image diffusion models as a teacher signal and combines this with an adversarial loss to ensure high image fidelity even in the low-step regime of one or two sampling steps. The text encoders of this model are OpenCLIP-ViT/G and CLIP-ViT/L. 

\item \noindent\textbf{Stable Diffusion 3.5 Medium:} This model is a Multimodal Diffusion Transformer with improvements (MMDiT-X) text-to-image model that features improved performance in image quality, typography, complex prompt understanding, and resource-efficiency. The text encoders of this model are OpenCLIP-ViT/G, CLIP-ViT/L and  T5-xxl. 

\item \noindent\textbf{FLUX.1 dev:} his model is built upon a 12-billion-parameter \textit{rectified flow Transformer} architecture, 
a diffusion-based framework reformulated as a continuous flow matching process for improved stability and efficiency. 
The model employs a dual text-encoder design (T5-xxl and CLIP-ViT/L) for robust semantic conditioning and integrates 
a high-capacity autoencoder for latent-space compression. 
\end{itemize}

\section{More Details of Baselines}\label{sec:baselines}
\begin{table}[h]
\scriptsize
\centering
\caption{Comparison of existing prompt stealing and prompt inversion Methods.}
\label{tab: baselines_methods}
\begin{tabular}{lccc}
\toprule
\textbf{Method}  & \textbf{Modifier} & \textbf{Optimization} & \textbf{Black-box} \\
\midrule
PEZ~\cite{10.5555/3666122.3668341}   & \XSolidBrush & \Checkmark & \XSolidBrush \\
PH2P~\cite{ph2p2024cvpr}                  & \XSolidBrush & \Checkmark & \XSolidBrush \\
BLIP~\cite{li2022blip}                 & \XSolidBrush & \XSolidBrush & \Checkmark  \\
CLIP-IG~\cite{clip_interrogator_github}             & \Checkmark & \XSolidBrush & \Checkmark \\
VGD~\cite{kim2025visuallyguideddecodinggradientfree}  & \XSolidBrush & \XSolidBrush & \Checkmark\\
PromptStealer~\cite{shen2024promptstealingattackstexttoimage}  & \Checkmark & \XSolidBrush & \Checkmark\\
\textbf{\sys~(Ours)}                      & \Checkmark & \Checkmark & \Checkmark\\

\bottomrule
\end{tabular}
\end{table}

We compare our approach with state-of-the-art prompt inversion and prompt stealing methods as summarized in~\autoref{tab: baselines_methods}. In addition, we include auxiliary baselines that derive prompts from image captioning and from vision–language models (VLMs).
\begin{itemize}
    \item \noindent\textbf{BLIP~\cite{https://doi.org/10.48550/arxiv.2201.12086}:} This is an image captioning model that learns from image-caption pairs~\cite{10.1007/978-3-319-10602-1_48} to connect vision and language.
    \item \noindent\textbf{CLIP Interrogator~\cite{clip_interrogator_github}:} This is a tool that uses CLIP and BLIP together to analyze an image and suggest text prompts that best describe it. In addition, it chooses extra descriptive modifiers from a predefined large-scale modifier pool.
    \item \noindent\textbf{VLM-as-an-expert:}   The vision-language model provides strong multimodal reasoning ability, allowing it to interpret complex visual content, align it with linguistic context, and generate accurate, coherent prompts. Here we use GPT-4o~\cite{hurst2024gpt} for this purpose.

    \item \noindent\textbf{PromptStealer~\cite{shen2024promptstealingattackstexttoimage}:} This is a prompt stealing attack that attempts to recover the prompt. It consists of a subject generator to infer the main subject and a modifier detector to identify descriptive modifiers. 
    \item \noindent\textbf{PH2P~\cite{ph2p2024cvpr}:} This method focuses on the later, more noisy timesteps and uses delayed projection into the discrete token vocabulary to recover human-readable prompts that reflect the visual content.  
    \item \noindent\textbf{VGD~\cite{kim2025visuallyguideddecodinggradientfree}:} This is a gradient-free hard prompt inversion method that uses a large language model and CLIP guidance to generate human-readable prompts aligned with visual content, without extra training.  
\end{itemize}

\section{More Details of Datasets}\label{sec:datasets}
We empirically evaluate our prompt stealing pipeline on four widely used datasets: MS COCO~\cite{10.1007/978-3-319-10602-1_48}, Flickr~\cite{young2014image}, Lexica~\cite{SQBZ24}, and DiffusionDB~\cite{wangDiffusionDBLargescalePrompt2022}. 
For the first three datasets, we randomly select 50 prompts and use the target text-to-image generative models to synthesize the corresponding images. 
For DiffusionDB, we treat it as an \emph{in-the-wild} dataset: instead of generating new images with the target models, we directly use 50 original images from the dataset, where the underlying generative models are unknown.

\begin{itemize}
    \item \noindent\textbf{MS COCO and Flickr:}
    MS COCO is a large benchmark of everyday scenes with natural photographs. Each image has multiple human-written captions. The captions are full sentences that name visible objects, actions, and simple scene context in plain language. Flickr is a captioning dataset of real-world photos collected from Flickr with five human-written sentences per image. These two datasets use sentence-style prompts that describe entities, actions, and scene layout without art-style tags. 
    \item \noindent\textbf{Lexica and DiffusionDB:}
    Lexica is a collection of user-entered prompts from an online prompt search service for text-to-image models. Prompts are keyword-focused and often include long chains of style, medium, camera, lighting, and quality terms; some entries also include negative phrases.
    DiffusionDB is the first large-scale text-to-image prompt dataset. It contains 14 million images generated by Stable Diffusion using prompts and hyperparameters specified by real users. Prompts are typically keyword lists with stylistic and technical modifiers rather than full sentences about photographic scenes. Prompts of these two datasets follow a \textit{“subject, modifiers”} pattern, where modifiers are comma-separated terms for style, medium, rendering setup, lighting, and quality.
\end{itemize}

\section{More Details of Metrics}
\label{sec:metrics}
We employ both image similarity and textual alignment metrics to quantitatively evaluate the quality of generated images and the stolen prompts. Specifically, we use CLIP Similarity and LPIPS Similarity for image comparison, and BERTScore and SBERT for textual alignment assessment.

\begin{itemize}[leftmargin=*]
    \item \noindent\textbf{CLIP Similarity:} 
    This metric measures the semantic similarity between the generated image $\hat{x}$ and the target image $x$. 
    It leverages the image encoder $f_\text{img}$ from CLIP~\cite{radford2021learning} to compute cosine similarity in the joint vision-language embedding space:
    \begin{equation}
        CLIP(\hat{x}, x) = 
        \frac{f_\text{img}(\hat{x}) \cdot f_\text{img}(x)}
        {\|f_\text{img}(\hat{x})\|\|f_\text{img}(x)\|},
    \end{equation}
    where a higher value indicates stronger semantic alignment between $\hat{x}$ and $x$.

    \item \noindent\textbf{LPIPS Similarity:}
    The Learned Perceptual Image Patch Similarity (LPIPS)~\cite{Zhang_2018_CVPR} metric evaluates perceptual similarity between two images by comparing deep features extracted from a pretrained convolutional neural network (e.g., AlexNet or VGG). 
    LPIPS computes the L2 distance between normalized features across multiple layers, reflecting human perceptual judgments of image quality. 
    A lower LPIPS score indicates higher perceptual similarity.

   \item \noindent\textbf{BERTScore:}
    To measure token-level textual alignment, we use BERTScore~\cite{zhang2020bertscoreevaluatingtextgeneration} with a token encoder $f_{\text{bert}}$.
    For candidate tokens $\{\hat{p}_i\}_{i=1}^{|\hat{p}|}$ and reference tokens $\{p_j\}_{j=1}^{|p|}$, define cosine similarity
    $c_{ij}=\frac{f_{\text{bert}}(\hat{p}_i)\cdot f_{\text{bert}}(p_j)}
    {\|f_{\text{bert}}(\hat{p}_i)\|\,\|f_{\text{bert}}(p_j)\|}$.
    Precision and recall are
    \begin{equation}
        \text{P}=\frac{1}{|\hat{p}|}\sum_{i}\max_{j} c_{ij}, \qquad
        \text{R} =\frac{1}{|p|}\sum_{j}\max_{i} c_{ij},
    \end{equation}
    and the final BERTScore is the F1:
    \begin{equation}
        \mathrm{BERTScore}(\hat{p}, p)=\frac{2 \text{P}\text{R}}{ \text{P}+\text{R}}.
    \end{equation}

    \item \noindent\textbf{SBERT:}
    Sentence-level textual alignment is computed with Sentence-BERT~\cite{reimers2019sentencebertsentenceembeddingsusing} using a sentence encoder $f_{\text{sbert}}$.
    Given a generated prompt $\hat{p}$ and a target prompt $p$, we take the cosine similarity of their sentence embeddings:
    \begin{equation}
        \mathrm{SBERT}(\hat{p}, p)=
        \frac{f_{\text{sbert}}(\hat{p}) \cdot f_{\text{sbert}}(p)}
             {\|f_{\text{sbert}}(\hat{p})\| \, \|f_{\text{sbert}}(p)\|},
    \end{equation}
    where a higher score indicates stronger semantic equivalence.
\end{itemize}

\section{More details of User Study}
\label{sec:user_study_dist}
\begin{figure}[ht]
  \centering
  \begin{subfigure}{0.48\linewidth}
    \includegraphics[width=\linewidth]{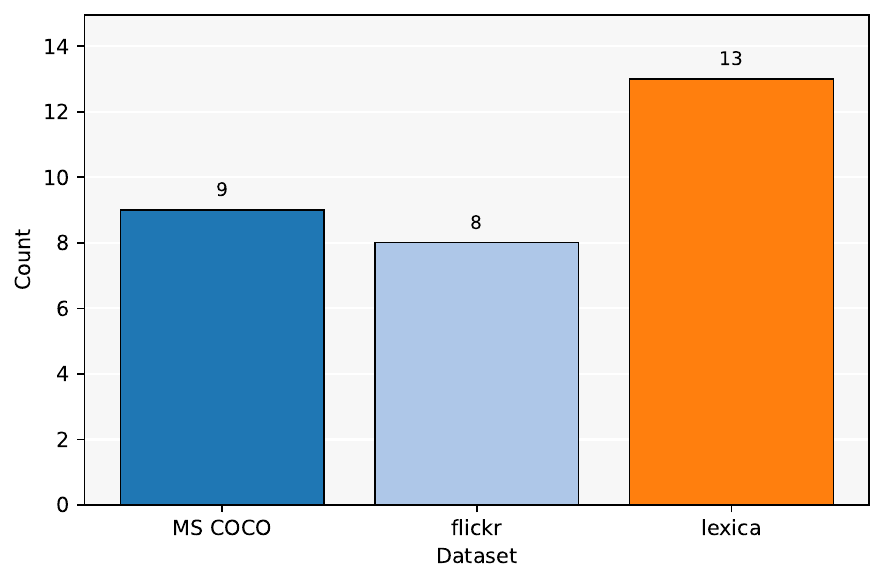}
    \caption{Samples per dataset}
    \label{fig:dist-dataset}
  \end{subfigure}\hfill
  \begin{subfigure}{0.48\linewidth}
    \includegraphics[width=\linewidth]{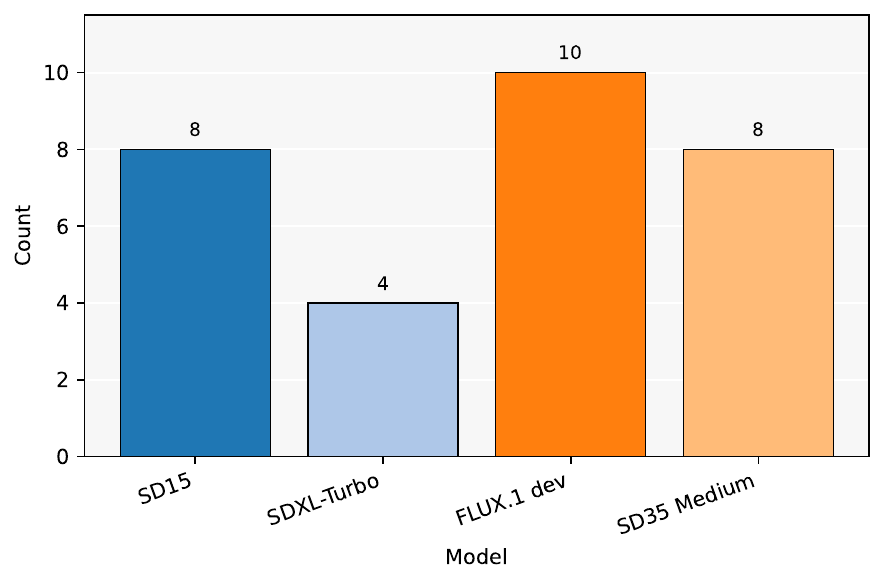}
    \caption{Samples per model}
    \label{fig:dist-model}
  \end{subfigure}
  \caption{Distributions of the sampled test cases in the user study. \textbf{SD15} is Stable Diffusion v1.5, and \textbf{SD35 Medium} means Stable Diffusion 3.5 Medium.}
  \label{fig:dists}
\end{figure}

We conduct a user study to assess the perceptual quality and fidelity of prompts recovered by different methods. We randomly select 30 image sets from three datasets and four generative models. Each set includes one target image and seven generated images produced by different prompt stealing or prompt inversion methods. During evaluation, participants are blinded to both the methods and the corresponding prompts, and the order of images within each set is randomly shuffled.

Participants rate the similarity between each generated image and its target on a 6-point Likert scale from 0 to 5, where 0 indicates \emph{not similar at all} and 5 indicates \emph{almost identical}. We compute the final score for each method as the average rating across all related images. The distribution of sampled test cases is illustrated in ~\autoref{fig:dists}.

\section{More details of Defenses}
\label{sec:defenses}
\begin{figure*}[h] 
  \centering
  \includegraphics[width=1.0\linewidth]{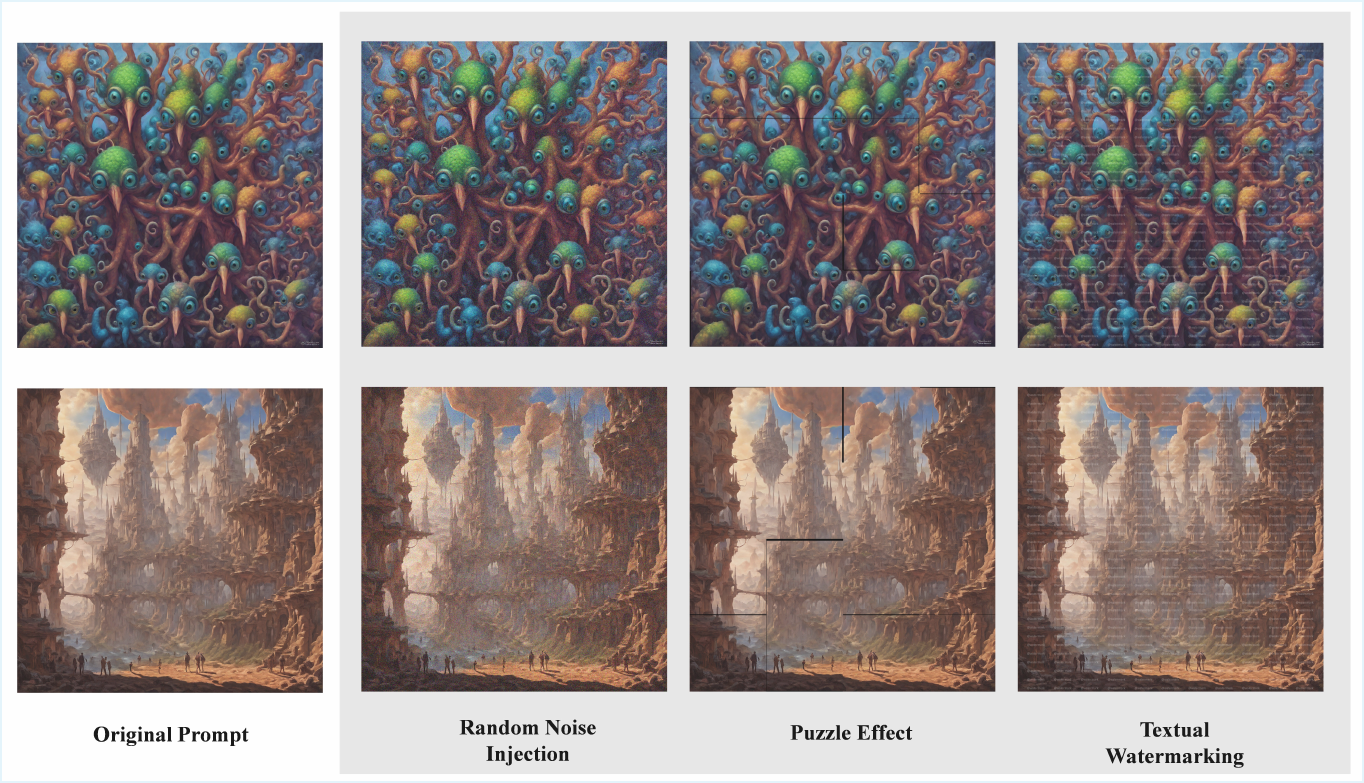} 
  \caption{Visualization of images after applying three representative defenses.}
  \label{fig:defenses}
\vspace{0.2cm}  
\end{figure*}
As shown in~\autoref{fig:defenses}, we investigate three representative defenses: (1) \textbf{Random noise Injection}, where Gaussian noise 
with a mean of 0 and a standard deviation of approximately 25 is added to subtly perturb pixel-level information; (2) \textbf{Puzzle effect}, which divides the image into a $4\times4$ grid and applies random local translations with a variability parameter of 3, 
which is applied in practice~\cite{Dianamoran@PromptBase}; and (3) \textbf{Textual watermarking}, which overlays a visible pattern such as “@watermark” across the image 
with a font size of 20 and row/column spacing of 20/30 pixels.

\section{More details of VLM-Based Mutators}
\label{sec:vlm_mutator_prompts}
To enable structured, targeted prompt refinement, we design a suite of \emph{VLM-based mutators} that operate on both subjects and modifiers. Each mutator leverages the vision-language reasoning ability of a powerful VLM (e.g., Qwen2-VL-2B-Instruct~\cite{Qwen2VL}) to generate linguistically natural and visually grounded edits.
Specifically, these mutators either paraphrase or enrich the subject for clarity and detail, or modify the descriptive and stylistic aspects of the prompt to enhance expressiveness and visual fidelity.
Below, we outline the five mutator types and their respective functions.

\begin{itemize}
    \item \underline{\textit{Subject-Paraphrase:}} Designed to rewrite the subject without altering its semantic meaning, this operator restructures the sentence to achieve higher linguistic diversity and naturalness while maintaining content fidelity to the original description.

    \item \underline{\textit{Subject-Enrich:}} This operator augments the subject by inserting concise, image-grounded details such as color, quantity, or pose. The enrichment strengthens the visual grounding of the prompt while preserving its syntactic structure and readability.


     \item \underline{\textit{Modifier-Generate:}} Designed to synthesize \textbf{a total new \emph{description} and \emph{style}} jointly from the image and the subject, this operator produces a compact, comma-separated tag string for scene facts together with a short style tag, ensuring strong visual grounding and aesthetic control.

    \item \underline{\textit{Modifier-Description:}} Designed to enhance descriptive richness, this operator refines or extends the \textbf{existing prompt’s description} by incorporating spatial relations, compositional layouts, and lighting attributes observed in the image. It effectively bridges literal captions and visually rich scene descriptions.

    \item \underline{\textit{Modifier-Style:}} Starting from the \textbf{current prompt's style}, this operator introduces or adjusts stylistic tokens such as medium, texture, camera lens, or rendering quality. It allows the prompt to better control the generative behavior of the text-to-image model and produce outputs with stronger artistic fidelity.
\end{itemize}

Besides these five primary operators, we also include an auxiliary mutator, \underline{\emph{Subject-Fix-Grammar}}, to remedy artifacts introduced by RL optimization (e.g., token repetition, minor grammatical errors, punctuation/spacing issues). This lightweight cleanup pass preserves the original semantics and word order while improving readability and coherence. We show the exact prompts for each mutation below.
\subsection{Shared System Prompt}
\begin{promptbox}{System Prompt (shared across mutators)}
you are a prompt mutator for text-to-image diffusion models.\\
given a base prompt and an input image, you must return EXACTLY ONE SINGLE-LINE JSON object.\\
- lowercase english only; no markdown; no code fences; no trailing commas; no extra text.\\
- never insert line breaks inside values.\\
- if the base prompt conflicts with the image, trust the image.\\
- be concrete and visual; do not invent invisible objects.\\
- field-specific rules:\\
  description: 15-35 words, write as a compact comma-separated tag string (not full sentences). include: subject and key attributes or pose, setting/location, composition/shot/angle, lighting, overall color tendency, one brief quality token (e.g., highly detailed or sharp focus), optional material/texture cue, artist, plus up to 2 simple negatives (e.g., no watermark, no text). use only visible facts.\\
  style: <= 12 words, a short comma-separated tag string of medium/movement/lens/quality only (e.g., digital painting, photorealistic, vector art, isometric, 35mm lens, 85mm lens, film grain, clean render). do not include scene facts or lighting.\\
  base\_prompt: <= 15 words, preserve the original meaning, clearer phrasing, avoid style or lighting tokens.\\
examples:\\
 input(base): 'two samurai duel in a bamboo forest'\\
 output(desc+style): {"description":"bamboo grove, two samurai facing between tall stalks, medium shot, eye-level, dappled sunlight, green tones, foreground leaves, no text","style":"ink illustration, ukiyo-e inspired, paper texture"}\\
 input(base): 'astronaut and robot on mars at dawn'\\
 output(modify-desc): {"description":"red dunes, astronaut left of small robot, wide shot, low angle, soft dawn light, cool shadows, distant mountains, no watermark"}\\
 input(base): 'portrait of an old musician in neon city'\\
 output(modify-style): {"style":"photorealistic, 85mm lens, cinematic still"}\\
 input(base): 'a child reading a book under a tree'\\
 output(paraphrase-base): {"base\_prompt":"a child reading beneath a tree"}\\
\end{promptbox}

\subsection{Subject Mutators}
\begin{promptbox}{Subject-Enrich (user prompt)}
task: ENRICH the base prompt by inserting concise, image-grounded modifiers WITHOUT changing its syntactic skeleton or word order.\\
base prompt: '{base\_prompt}'\\
you will receive the IMAGE together with this message. use ONLY details that are VISIBLE in the image.\\
output schema (single line json): {"base\_prompt":"..."}\\
constraints:\\
- preserve the original tokens as an ordered subsequence: do not delete, replace, or reorder existing words; no synonym substitution.\\
- keep the subject–verb–object–prepositional structure intact.\\
- INSERT 2–6 words total, placed immediately AFTER the nouns/verbs they modify (adjectives/appositives for nouns; short adverbs/adjuncts for verbs).\\
- allowed insertions: count/quantity (one/two), object attributes (color, size, material), pose/state, simple spatial cues relative to BACKGROUND (e.g., near the fence, in front of the gate), and other concrete scene facts visible in the image.\\
- forbid insertions about style/lighting/lens/artist or abstract aesthetics (these belong to style).\\
- if a detail is uncertain or not visible, DO NOT add it; trust the image over the text.\\
- lowercase only; no quotes; no extra commentary.\\
examples:\\
 base: 'a child reading a book under a tree'\\
 enriched: {'base\_prompt':'a small child quietly reading a worn book under a shady tree'}\\
 base: 'a dog runs across a field'\\
 enriched: {'base\_prompt':'a brown dog runs swiftly across a grassy field'}\\
return ONLY the json line.\\
\end{promptbox}

\begin{promptbox}{ Subject-Fix-Grammar (user prompt)}
task: CLEANUP the base prompt by fixing grammar/spelling, removing duplicated words/phrases, and correcting spacing/punctuation ONLY.\\
base prompt: '{base\_prompt}'\\
output schema (single line json): {"base\_prompt":"..."}\\
constraints:\\
- do NOT add any new content or modifiers; do NOT introduce synonyms; do NOT reorder clauses.\\
- preserve the original subject–verb–object–prepositional order and overall sentence structure.\\
- only remove repeated tokens/phrases, fix typos, collapse multiple spaces, and standardize minimal punctuation.\\
- keep length approximately unchanged (within ±2 words of the original).\\
- lowercase only; no quotes; no extra commentary.\\
return ONLY the json line.\\
\end{promptbox}

\begin{promptbox}{Subject-Paraphrase (user prompt)}
task: PARAPHRASE the base prompt WITHOUT changing its meaning, and enforce the structure:\\
WHO/WHAT + is doing + WHERE.\\
base prompt: '{base\_prompt}'\\
output schema (single line json): {"base\_prompt":"..."}\\
constraints:\\
- <= 15 words total.\\
- structure must be strictly: <who/what> + 'is/are' + <present participle> + <where-phrase>.\\
- examples: 'a child is reading under a tree'; 'two samurai are dueling in a bamboo forest'; 'a red car is driving along a rainy street'.\\
- no style/lighting/lens/artist tokens.\\
- preserve entities and relations; trust the image if there is a conflict.\\
- lowercase only; no quotes; no extra commentary.\\\\
return ONLY the json line.\\
\end{promptbox}

\subsection{Modifier-Generate}
\begin{promptbox}{GEN\_DESC\_STYLE (user prompt)}
task: generate a NEW description and a NEW style from the base prompt and the image.
base prompt: '{base\_prompt}'\\
output schema (single line json): {"description":"...","style":"..."}\\
constraints:\\
- description: 15–35 words, compact comma-separated tag string (not full sentences).\\
- include, in order when possible: subject+attributes/pose, setting, composition/shot/angle, lighting, overall color tendency, one quality token, optional material/texture, artist, up to 2 simple negatives.\\
- include one explicit spatial or depth cue (e.g., foreground, in front of distant hills).\\
- style: <= 12 words; medium/movement/lens/quality only; no scene facts; no lighting.\\
return ONLY the json line.\\
\end{promptbox}

\begin{promptbox}{Modifier-Style (user prompt)}
task: CHANGE STYLE ONLY while preserving entities and relations in the current description.\\
current style: {style-from-parent}\\
current description: {description-from-parent}\\
base prompt: '{base\_prompt}'\\
output schema (single line json): {"style":"..."}\\
constraints:\\
- concise comma-separated tag string (<=12 words) of medium/movement/lens/quality.\\
- do NOT include scene facts or lighting.\\
- keep the aesthetic consistent; improve clarity or fidelity.\\
return ONLY the json line.\\
\end{promptbox}

\begin{promptbox}{Modifier-Description (user prompt)}
task: CHANGE DESCRIPTION ONLY while preserving the subject meaning and current style.\\
current description: {description-from-parent}\\
current style: {style-from-parent}\\
base prompt: '{base\_prompt}'\\
output schema (single line json): {"description":"..."}\\
constraints:\\
- compact comma-separated tag string (15–35 words).\\
- include: subject+attributes/pose, setting, composition/shot/angle, lighting, overall color tendency, one quality token, optional material/texture, artist, up to 2 negatives.\\
- include one explicit spatial or depth cue.\\
- concrete visible details only; avoid abstractions; do not change the style semantics.\\
return ONLY the json line.\\
\end{promptbox}

\section{Token-Level Textual Alignment}\label{sec:token_alignment}
\begin{table*}[h]
\centering
\caption{Token-level textual alignment comparison across datasets.}
\resizebox{\linewidth}{!}{
\begin{tabular}{ll*{12}{p{0.9cm}}}
\toprule
\multirow{2}{*}{Dataset} & \multirow{2}{*}{Method} & 
\multicolumn{3}{c}{Stable Diffusion v1.5} & 
\multicolumn{3}{c}{SDXL-Turbo} & 
\multicolumn{3}{c}{FLUX.1 dev} & 
\multicolumn{3}{c}{Stable Diffusion 3.5 Medium} \\
\cmidrule(lr){3-5} \cmidrule(lr){6-8} \cmidrule(lr){9-11} \cmidrule(lr){12-14}
 & & P$\uparrow$ & R$\uparrow$ & F1$\uparrow$ & P$\uparrow$ & R$\uparrow$ & F1$\uparrow$ & P$\uparrow$ & R$\uparrow$ & F1$\uparrow$ & P$\uparrow$ & R$\uparrow$ & F1$\uparrow$ \\
\midrule
\multirow{7}{*}{MS COCO} 
 & BLIP            &0.905&0.911&0.908 &0.905&0.912&0.909 &0.907&0.917&0.912 &0.912&0.921&0.916 \\
 & CLIP-IG         &0.803&0.894&0.846 &0.793&0.896&0.841 &0.797&0.898&0.844 &0.797&0.900&0.845 \\
 & VLM-as-expert      &0.826&0.898&0.860 &0.826&0.898&0.861 &0.826&0.904&0.863 &0.831&0.908&0.868 \\
 & PH2P            &0.761&0.827&0.793 &0.768&0.827&0.797 &0.765&0.827&0.795 &0.757&0.827&0.790 \\
 & PromptStealer   &0.905&0.919&0.912 &0.915&0.922&0.918 &0.911&0.921&0.916 &0.917&0.926&0.922 \\
 & VGD             &0.822&0.874&0.847 &0.829&0.880&0.853 &0.827&0.874&0.850 &0.822&0.874&0.847 \\
 & \cellcolor{blue!10}\sys~(Ours) 
   & \cellcolor{blue!10}0.904 & \cellcolor{blue!10}0.901 & \cellcolor{blue!10}0.902 
   & \cellcolor{blue!10}0.903 & \cellcolor{blue!10}0.906 & \cellcolor{blue!10}0.904 
   & \cellcolor{blue!10}0.903 & \cellcolor{blue!10}0.919 & \cellcolor{blue!10}0.911 
   & \cellcolor{blue!10}0.904 & \cellcolor{blue!10}0.915 & \cellcolor{blue!10}0.909 \\
\midrule
\multirow{7}{*}{Flickr} 
 & BLIP            &0.901&0.887&0.894 &0.907&0.888&0.897 &0.901&0.891&0.896 &0.903&0.889&0.896 \\
 & CLIP-IG         &0.809&0.876&0.841 &0.800&0.878&0.837 &0.803&0.880&0.839 &0.802&0.880&0.839 \\
 & VLM-as-expert      &0.835&0.890&0.861 &0.832&0.887&0.859 &0.838&0.900&0.867 &0.840&0.901&0.869 \\
 & PH2P            &0.767&0.814&0.789 &0.762&0.813&0.786 &0.758&0.813&0.784 &0.763&0.813&0.787 \\
 & PromptStealer   &0.895&0.889&0.892 &0.911&0.898&0.904 &0.910&0.895&0.902 &0.911&0.899&0.905 \\
 & VGD             &0.827&0.858&0.842 &0.826&0.860&0.842 &0.824&0.856&0.839 &0.824&0.856&0.839 \\
 & \cellcolor{blue!10}\sys~(Ours) 
   & \cellcolor{blue!10}0.906 & \cellcolor{blue!10}0.904 & \cellcolor{blue!10}0.905 
   & \cellcolor{blue!10}0.917 & \cellcolor{blue!10}0.907 & \cellcolor{blue!10}0.912 
   & \cellcolor{blue!10}0.914 & \cellcolor{blue!10}0.901 & \cellcolor{blue!10}0.907 
   & \cellcolor{blue!10}0.913 & \cellcolor{blue!10}0.905 & \cellcolor{blue!10}0.909 \\
\midrule
\multirow{7}{*}{Lexica} 
 & BLIP            &0.855&0.783&0.817 &0.855&0.782&0.816 &0.853&0.782&0.815 &0.856&0.784&0.818 \\
 & CLIP-IG         &0.828&0.823&0.825 &0.827&0.822&0.825 &0.827&0.824&0.826 &0.822&0.825&0.823 \\
 & VLM-as-expert      &0.822&0.802&0.812 &0.825&0.803&0.814 &0.823&0.803&0.813 &0.826&0.806&0.816 \\
 & PH2P            &0.761&0.774&0.767 &0.769&0.775&0.772 &0.761&0.774&0.768 &0.760&0.775&0.767 \\
 & PromptStealer &0.865&0.820&0.842 &0.869&0.817&0.842 &0.871&0.818&0.843 &0.872&0.819&0.844 \\
 & VGD             &0.822&0.791&0.806 &0.825&0.790&0.807 &0.790&0.816&0.803 &0.814&0.789&0.801 \\
 & \cellcolor{blue!10}\sys~(Ours)
   & \cellcolor{blue!10}0.844 & \cellcolor{blue!10}0.810 & \cellcolor{blue!10}0.826
   & \cellcolor{blue!10}0.846 & \cellcolor{blue!10}0.809 & \cellcolor{blue!10}0.827
   & \cellcolor{blue!10}0.850 & \cellcolor{blue!10}0.820 & \cellcolor{blue!10}0.835
   & \cellcolor{blue!10}0.856 & \cellcolor{blue!10}0.823 & \cellcolor{blue!10}0.839 \\
\bottomrule
\end{tabular}}
\label{tab: prompt_align_baselines}
\end{table*}

\begin{table}[h]
\centering
\caption{Token-level textual alignment comparison on \textit{in-the-wild} datasets with SDXL-Turbo.}
\scriptsize
\begin{tabular}{lccc}
\toprule
\multirow{2}{*}{Method} & 
\multicolumn{3}{c}{\textbf{Textual Alignment}} \\
\cmidrule(lr){2-4}
 & P$\uparrow$ & R$\uparrow$ & F1$\uparrow$ \\
\midrule
BLIP            &0.857  &0.815  &0.835  \\
CLIP-IG         &0.819  &0.841  &0.830  \\
VLM-as-expert     &0.820  &0.823  &0.821  \\
PH2P            &0.770  &0.791  &0.780  \\
PromptStealer   &0.860  &0.841  &0.850  \\
VGD             &0.814  &0.806  &0.810  \\
\cellcolor{blue!10}\sys~(Ours) 
   &\cellcolor{blue!10}0.838 & \cellcolor{blue!10}0.825 & \cellcolor{blue!10}0.821\\
\bottomrule
\end{tabular}
\label{tab:inwild_prf1}
\end{table}

We evaluate token-level textual alignment on MS COCO, Flickr, and Lexica using four T2I backbones (Stable Diffusion v1.5, SDXL-Turbo, FLUX.1 dev, and Stable Diffusion 3.5 Medium). As shown in \autoref{tab: prompt_align_baselines}, \sys delivers the highest or near-highest scores on Precision(P), Recall(R), and F1 Score(F1) across datasets and models, indicating that our recovered prompts are both precise and comprehensive with respect to image content. We attribute these gains to the two-stage design: (i) \emph{RL-Based Prompt Inversion} improves subject fidelity and structural consistency; (ii) \emph{Fuzz Testing–Powered Prompt Optimization} introduces stylistic and compositional modifiers without sacrificing semantic grounding. Notably on Lexica, where prompts contain diverse, artist-crafted modifiers, \sys maintains strong F1, demonstrating robustness to varied prompt distributions.

\paragraph{In-the-Wild Setting.} We further assess generalization on images whose source generators are unknown. We use SDXL-Turbo as the text-to-image generative model. 
As reported in \autoref{tab:inwild_prf1}, \sys attains competitive textual alignment, demonstrating that our black-box, semantics-driven optimization transfers effectively to unconstrained scenarios.

\section{Ablation Study}
\label{sec:app_ablation}

\begin{table}[h]
\centering
\caption{Impact of different VLMs used as mutators in the Fuzz Testing–Powered Optimization stage.}
\scriptsize
\begin{tabular}{lccc}
\toprule
\multirow{2}{*}{Mutator Model} & 
\multicolumn{2}{c}{\textbf{Image Similarity}} & 
\multicolumn{1}{c}{\textbf{Textual Alignment}} \\
\cmidrule(lr){2-3}\cmidrule(lr){4-4}
 & CLIP$\uparrow$ & LPIPS$\downarrow$ & SBERT$\uparrow$ \\
\midrule

Qwen2-VL-2B-Instruct &0.911  &0.420  &0.591  \\
Qwen2-VL-7B-Instruct &0.910  &0.447  &0.597  \\
GPT-4o &0.917  &0.438  &0.568  \\
\bottomrule
\end{tabular}
\label{tab:mutator_impact}
\end{table}

\noindent \textbf{Impact of Different Mutators.} We investigate the impact of using different vision–language models (VLMs) as mutators in the Fuzz Testing–Powered Optimization stage. Specifically, we select three representative VLMs with varying model capacities and architectures: \textbf{Qwen2-VL-2B-Instruct}, \textbf{Qwen2-VL-7B-Instruct}, and \textbf{GPT-4o}. The Qwen2-VL family represents open-source VLMs with strong multimodal grounding and scalable size variants, while GPT-4o serves as a powerful proprietary model that excels in multimodal reasoning and image–text alignment. 
As shown in \autoref{tab:mutator_impact}, all three models enable effective prompt refinement, demonstrating the general applicability of our fuzz-testing optimization framework. 
Smaller open-source models such as Qwen2-VL-2B-Instruct already achieve competitive performance, indicating that our method does not rely on large-parameter models or extensive computational resources to generate high-quality modifiers and achieve strong inversion performance.

\begin{table}[h]
\centering
\caption{Impact of \textit{Phase~I} (RL-based Prompt Inversion) and \textit{Phase~II} (Fuzz testing-Powered Optimization). }
\scriptsize
\begin{tabular}{llccc}
\toprule
\multirow{2}{*}{Dataset} & \multirow{2}{*}{Method} & 
\multicolumn{2}{c}{\textbf{Image Similarity}} & 
\multicolumn{1}{c}{\textbf{Textual Alignment}} \\
\cmidrule(lr){3-4} \cmidrule(lr){5-5}
 &  & CLIP$\uparrow$ & LPIPS$\downarrow$ & SBERT$\uparrow$ \\
\midrule
\multirow{4}{*}{Flickr}
 & CLIP-IG                      & 0.835 & 0.478 & 0.468 \\
 & \textit{Phase~I} only        & 0.859 & 0.440 & 0.559 \\
 & \textit{Phase~II} only       & 0.891 & 0.433 & 0.541 \\
 & \textbf{\sys}                & 0.910 & 0.405 & 0.603 \\
\midrule
\multirow{4}{*}{Lexica}
 & CLIP-IG                      & 0.856 & 0.451 & 0.591 \\
 & \textit{Phase~I} only        & 0.832 & 0.475 & 0.440 \\
 & \textit{Phase~II} only       & 0.889 & 0.434 & 0.559 \\
 & \textbf{\sys}                & 0.911 & 0.420 & 0.591 \\
\bottomrule
\end{tabular}
\label{tab:ablation_phase_impact}
\end{table}

\noindent \textbf{Impact of the Two Phase Design.}
We assess each stage separately and together using the ablation in~\autoref{tab:ablation_phase_impact}. 
First, we validate \emph{Phase~I} by replacing its outputs with BLIP~\cite{li2022blip} captions that are then used to initialize \emph{Phase~II}. 
Second, we disable \emph{Phase~II} and directly evaluate prompts from \emph{Phase~I}. 
On Flickr dataset, the prompts mainly describe the subject. 
\emph{Phase~I} already surpasses the best baselines in both image similarity and semantic alignment, and adding \emph{Phase~II} yields additional improvements that produce the best overall results. 
On Lexica dataset, the prompts include the subject and diverse modifiers. 
\emph{Phase~I} alone is not sufficient, while \emph{Phase~II} brings clear gains by exploring and refining modifiers. 
Using only \emph{Phase~II} with BLIP initialization improves over the baseline but remains below the full two-phase system. 
These results show that the two stages are complementary and both are necessary, consistent with our query budget analysis.
\emph{Phase~I} provides strong subject recovery that offers a high quality initialization for \emph{Phase~II} and enables the second stage to converge to a higher final value. 
\emph{Phase~II} in turn refines the subject inferred by \emph{Phase~I} and systematically augments it with appropriate modifiers, leading to the highest image similarity and semantic alignment across datasets.

\section{Implementation Details}
\label{sec:implementation}

\subsection{RL-Based Prompt Inversion}
During imitation learning (IL), we freeze the BLIP backbone and train a lightweight adapter MLP to imitate expert next-token decisions. Each expert trajectory consists of $(h_t, y_{t+1})$ pairs, where $h_t$ is the decoder hidden state of the partial prompt and $y_{t+1}$ is the next ground-truth token sampled from BLIP's caption generation.
In practice, we directly use BLIP to generate $10$ expert trajectories per image.
We train the adapter using cross-entropy loss on the decoder logits. Training is conducted for $2000$ epochs with the Adam optimizer, a learning rate of $3\times10^{-4}$, and a batch size of $8$.
The input hidden dimension is $768$, and the adapter MLP expands it to $1536$ and projects it back to $768$ dimensions using a two-layer architecture with ReLU activation.

After imitation pretraining, we fine-tune the model using the Proximal Policy Optimization (PPO) algorithm within the same prompt-generation environment.
We initialize the actor and critic networks from the pretrained adapter checkpoint.
The PPO hyperparameters are discount factor $\gamma=0.98$ and clipping threshold $\epsilon=0.2$.
The actor and critic learning rates are $1\times10^{-4}$ and $5\times10^{-4}$, respectively.
We update the PPO agent every 150 environment steps, and perform $K=4$ epochs of policy and value updates for each batch of collected trajectories. We train for up to 100 total epochs. For the shaped reward, we set the scaling coefficient $\beta$ to 10. 

\subsection{Fuzz Testing-Powered Prompt Optimization}

We adopt a query-budgeted fuzz testing optimization to explore the prompt space efficiently.  
The total query budget is set to $100$ image generations per optimization run.  
During the first $30$ queries, only the subject-related part of the prompt is mutated to refine the core semantic content. 
In the remaining $70$ queries, both the subject and modifiers are jointly optimized to improve compositional richness and visual fidelity. 
For experiments on the MS~COCO and Flickr datasets, we restrict all 100 queries to optimizing the subject component.
The mutation process maintains a seed pool of $5$ candidate prompts, which are iteratively sampled, expanded, and replaced based on their CLIP Similarity until the query budget is exhausted.

\sys is implemented with Python 3.10 and PyTorch
2.6.0. We conducted all experiments on a Ubuntu 20.04 server
equipped with 1 A100 SXM4 GPU.

\section{Potential-Based Reward Shaping}
\label{sec:reward_shaping}

We provide a formal analysis of the \emph{potential-based reward shaping} mechanism used in the~\autoref{sec:method}. 
We first show that under mild assumptions, the potential-based transformation preserves the optimal policy of the original MDP. 
We then derive how such shaping influences the learning dynamics by improving the reward density, gradient variance, and initialization of value estimates.

\subsection{Setup and Definitions}

We recall the MDP $\mathcal{M} = (\mathcal{S}, \mathcal{A}, P, r, \gamma)$ defined in Section~\ref{sec:method}, where $\mathcal{S}$ denotes the state space, $\mathcal{A}$ the action space, $P(s' \mid s,a)$ the transition kernel, $r(s,a)$ the reward function, and $\gamma \in [0,1)$ the discount factor. 
For the sake of theoretical completeness, we slightly generalize the reward function here to the form $r(s,a,s')$, which makes explicit its possible dependence on both the action $a$ and the resulting next state $s'$. 
This modification is purely notational—conceptually equivalent to $r(s,a)$ used in ~\autoref{sec:method}—and serves to accommodate the upcoming definition of state-dependent shaping functions $F(s,a,s')$. 
Under this general form, the value and action-value functions of a policy $\pi$ are expressed as
\begin{align}
V_\pi(s) &= \mathbb{E}_\pi\!\left[ \sum_{t=0}^{\infty} \gamma^t \, r(s_t, a_t, s_{t+1}) \;\middle|\; s_0 = s \right], \\
Q_\pi(s,a) &= \mathbb{E}_\pi\!\left[ \sum_{t=0}^{\infty} \gamma^t \, r(s_t, a_t, s_{t+1}) \;\middle|\; s_0 = s,\, a_0 = a \right].
\end{align}

To introduce reward shaping, we define a bounded \emph{potential function} $\Phi: \mathcal{S} \rightarrow \mathbb{R}$, which assigns a scalar potential to each state, and construct the shaped reward
\begin{equation}
r'(s,a,s') = r(s,a,s') + F(s,a,s'), 
\quad \text{where} \quad 
F(s,a,s') = \gamma \Phi(s') - \Phi(s).
\end{equation}
Intuitively, the term $F(s,a,s')$ measures the potential difference between successive states, scaled by the discount factor. 
Because this term depends solely on the states and not on the agent’s policy, it modifies the learning dynamics without altering the underlying optimization objective, making it particularly suitable for theoretical analysis.

\subsection{Theorem: Policy Invariance under Potential-Based Shaping}

\begin{theorem}
Let $V_\pi$, $Q_\pi$ be the original value functions and $V'_\pi$, $Q'_\pi$ those under the shaped reward $r'$. 
Then for any bounded $\Phi$, the following holds:
\begin{align}
V'_\pi(s) &= V_\pi(s) - \Phi(s), \label{eq:vpi_invariance}\\
Q'_\pi(s,a) &= Q_\pi(s,a) - \Phi(s), \label{eq:qpi_invariance}
\end{align}
and therefore
\begin{equation}
A'_\pi(s,a) = Q'_\pi(s,a) - V'_\pi(s) = A_\pi(s,a),
\end{equation}
implying that $\arg\max_a Q'^*(s,a) = \arg\max_a Q^*(s,a)$, i.e., the optimal policy is preserved.
\end{theorem}

\paragraph{Proof.}
Starting from the shaped return:
\begin{align}
G'_t 
&= \sum_{k=0}^{\infty} \gamma^{k} \big[ r_{t+k} + \gamma \Phi(s_{t+k+1}) - \Phi(s_{t+k}) \big] \\
&= \underbrace{\sum_{k=0}^{\infty} \gamma^k r_{t+k}}_{G_t}
   + \sum_{k=0}^{\infty} \big( \gamma^{k+1} \Phi(s_{t+k+1}) - \gamma^{k} \Phi(s_{t+k}) \big).
\end{align}
The second summation is a telescoping series:
\begin{align}
\sum_{k=0}^{n} (\gamma^{k+1} \Phi_{k+1} - \gamma^{k} \Phi_{k}) 
= - \Phi(s_t) + \gamma^{n+1} \Phi(s_{t+n+1}),
\end{align}
and taking the limit $n\to\infty$, the terminal term vanishes because $\Phi$ is bounded and $\gamma<1$. 
Thus
\begin{equation}
G'_t = G_t - \Phi(s_t).
\end{equation}
Taking expectations under policy $\pi$ gives:
\begin{align}
V'_\pi(s) = \mathbb{E}_\pi[G'_t \mid s_t=s] = V_\pi(s) - \Phi(s),
\end{align}
which proves \eqref{eq:vpi_invariance}. 
For $Q'_\pi$, by the Bellman definition under $r'$:
\begin{align}
Q'_\pi(s,a) 
&= \mathbb{E}\!\left[ r'(s,a,s') + \gamma V'_\pi(s') \right] \\
&= \mathbb{E}\!\left[ r(s,a,s') + \gamma\Phi(s') - \Phi(s) + \gamma (V_\pi(s') - \Phi(s')) \right] \\
&= \mathbb{E}\!\left[ r(s,a,s') + \gamma V_\pi(s') \right] - \Phi(s) = Q_\pi(s,a) - \Phi(s),
\end{align}
proving \eqref{eq:qpi_invariance}.
Since subtracting $\Phi(s)$ does not depend on $a$, the greedy policy w.r.t.\ $Q'_\pi$ is identical to that of $Q_\pi$.

\subsection{Bellman Operator View and Corollary}

Let $\mathcal{T}_\pi$ and $\mathcal{T}'_\pi$ denote the Bellman operators:
\begin{align}
(\mathcal{T}_\pi V)(s) &= \mathbb{E}_{a\sim\pi}\!\left[ r(s,a,s') + \gamma V(s') \right], \\
(\mathcal{T}'_\pi V)(s) &= \mathbb{E}_{a\sim\pi}\!\left[ r'(s,a,s') + \gamma V(s') \right].
\end{align}
Then $\mathcal{T}'_\pi(V-\Phi) = \mathcal{T}_\pi(V) - \Phi$. 
Therefore, if $V_\pi$ is a fixed point of $\mathcal{T}_\pi$, then $V_\pi - \Phi$ is a fixed point of $\mathcal{T}'_\pi$. 
The same holds for the optimal Bellman operator $\mathcal{T}$, so $V'^* = V^* - \Phi$ and $Q'^* = Q^* - \Phi$. 
This formalizes that potential shaping merely shifts the value function manifold by a state-dependent bias but leaves the contraction and optimality conditions invariant.

\subsection{Potential-Based Reward Shaping under PPO Training}
\label{app:reward_shaping_ppo}

\paragraph{Proposition A.3 (PPO + GAE invariance under potential-based shaping).}
Let the shaped reward be
\begin{equation}
r'(s_t,a_t,s_{t+1}) \;=\; r(s_t,a_t,s_{t+1}) \;+\; \gamma\,\Phi(s_{t+1}) - \Phi(s_t),
\end{equation}
with the same discount factor $\gamma$ as the training objective.
PPO employs a critic $\hat V_\theta$ and uses generalized advantage estimation (GAE)
with temporal-difference residuals
\begin{equation}
\delta_t \;=\; r_t + \gamma\,\hat V_\theta(s_{t+1}) - \hat V_\theta(s_t), 
\qquad 
\widehat A_t \;=\; \sum_{l\ge 0}(\gamma\lambda)^l\,\delta_{t+l}.
\end{equation}
For the shaped MDP, define the \emph{shaped critic}
\begin{equation}
\hat V'_\theta(s) \;=\; \hat V_\theta(s) - \Phi(s),
\end{equation}
and compute
\begin{equation}
\delta'_t \;=\; r'_t + \gamma\,\hat V'_\theta(s_{t+1}) - \hat V'_\theta(s_t), 
\qquad
\widehat A'_t \;=\; \sum_{l\ge 0}(\gamma\lambda)^l\,\delta'_{t+l}.
\end{equation}

\textbf{Claim.}
For every trajectory and $\lambda\in[0,1]$,
\[
\delta'_t \equiv \delta_t, \qquad \widehat A'_t \equiv \widehat A_t.
\]

\textbf{Proof.}
Substitute $r'_t=r_t+\gamma\Phi(s_{t+1})-\Phi(s_t)$ and $\hat V'_\theta=\hat V_\theta-\Phi$:
\begin{align*}
\delta'_t
&= \underbrace{r_t+\gamma\Phi(s_{t+1})-\Phi(s_t)}_{r'_t}
+ \gamma(\hat V_\theta(s_{t+1})-\Phi(s_{t+1}))
- (\hat V_\theta(s_t)-\Phi(s_t)) \\
&= r_t + \gamma \hat V_\theta(s_{t+1}) - \hat V_\theta(s_t)
= \delta_t.
\end{align*}
Since GAE is a linear operator on the TD errors, 
$\widehat A'_t=\widehat A_t$ follows directly. 
\hfill$\square$

\paragraph{Corollary A.4 (PPO surrogate and value loss are unchanged).}
The PPO clipped surrogate objective
\begin{equation}
\mathcal{L}^{\text{CLIP}}(\theta)
= \mathbb{E}\!\left[\min\!\left(
r_t(\theta)\,\widehat A'_t,\;
\mathrm{clip}\big(r_t(\theta),1\!\pm\!\epsilon\big)\,\widehat A'_t
\right)\right],
\quad r_t(\theta)=\frac{\pi_\theta(a_t|s_t)}{\pi_{\theta_{\text{old}}}(a_t|s_t)},
\end{equation}
is identical to the unshaped objective since $\widehat A'_t\!\equiv\!\widehat A_t$.
The value regression loss is likewise invariant if both targets and predictions
are shifted consistently:
\begin{equation}
(R'_t - \hat V'_\theta(s_t))^2
= ((R_t-\Phi(s_t)) - (\hat V_\theta(s_t)-\Phi(s_t)))^2
= (R_t - \hat V_\theta(s_t))^2.
\end{equation}
Entropy regularization is unaffected. Hence, PPO updates on shaped data are
mathematically identical to unshaped updates when the critic is shifted by $-\Phi$.

\paragraph{Remark.}
This proposition is the PPO/GAE analogue of the classical policy-invariance theorem of Ng, Harada, and Russell~\cite{ng1999policy}.
Potential-based shaping leaves the true advantages $A'_\pi=A_\pi$ unchanged,
and with the critic shift $\hat V'=\hat V-\Phi$, it also leaves the \emph{estimated}
advantages $\widehat A'_t$ identical sample-by-sample.

\subsection{How Shaping Can Accelerate PPO Training Without Changing What Is Optimal}

If shaping is implemented exactly as above, the PPO actor and critic updates are
algebraically identical to those on the unshaped MDP.  Any speed-up therefore arises
from improved learning dynamics rather than a change in the underlying objective.
Below we summarize the PPO-specific mechanisms through which shaping can still help.

\paragraph{(1) Variance reduction via informed baselines.}
In actor–critic methods, any state-dependent baseline can be subtracted from returns
without biasing the gradient.  Choosing $\Phi$ correlated with returns provides an
additional control-variate that reduces the variance of $\widehat A_t$,
especially during early training when $\hat V_\theta$ is inaccurate.
This can be realized by using the shaped critic $\hat V'=\hat V-\Phi$,
or equivalently by adding $\Phi(s)$ to the baseline term in advantage computation.

\paragraph{(2) Finite-horizon credit assignment.}
PPO computes advantages over finite rollout segments.  When partial-episode bootstrapping
is used, the $\Phi$ terms cancel exactly (Proposition~A.3).  
If truncated episodes are instead treated as terminal, shaping introduces a boundary term,
providing denser feedback near segment boundaries and potentially improving early
credit assignment---while still leaving the optimal policy unchanged.

\paragraph{(3) Critic warm-start and architectural priors.}
Setting the initial critic to $\hat V_0(s)\!\approx\!\Phi(s)$ or embedding $\Phi(s)$
as a fixed residual in the value head supplies a useful prior.
Although the PPO gradients remain unbiased and unchanged in theory,
better initialization of the critic often stabilizes optimization and accelerates
the convergence of the actor–critic loop.

\paragraph{(4) Intentional partial shaping.}
Some implementations simply replace $r$ with $r'$ while keeping the critic unshifted.
Then $\widehat A'_t \!\neq\! \widehat A_t$ early on, producing progress-aligned
advantages that accelerate early learning.
While this breaks the exact gradient equivalence, the optimal policy is still
preserved by the theoretical invariance of potential-based shaping.

\section{More Qualitative Results}\label{sec:qualitative_results}

\begin{figure*}[h] 
  \centering
  \includegraphics[width=1.0\linewidth]{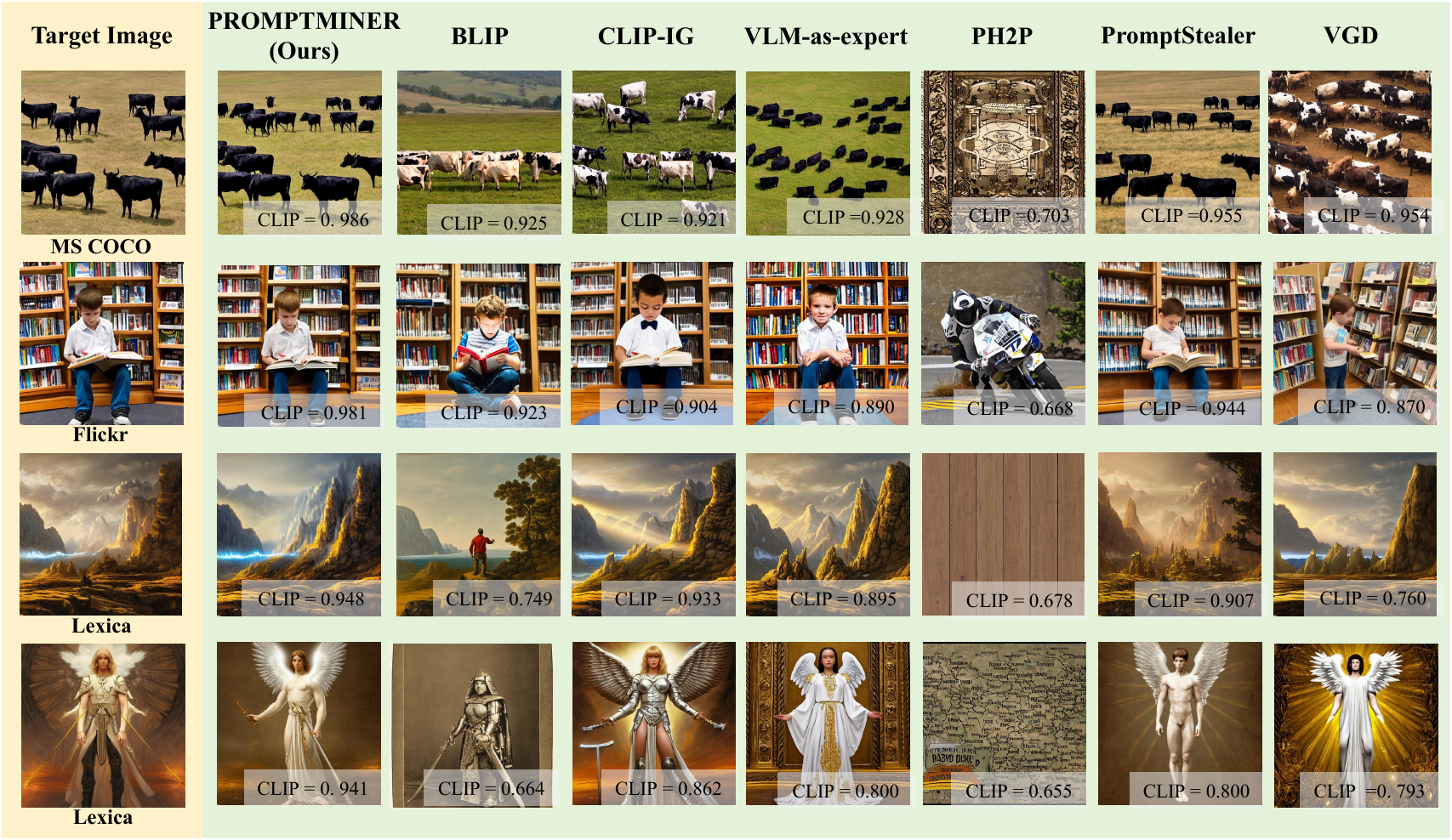} 
  \caption{Visualization of images generated by Stable Diffusion v1.5 compared with target image.}
  \label{fig:img_sim_visual_sd15}
\end{figure*}
\begin{figure*}[h] 
  \centering
  \includegraphics[width=1.0\linewidth]{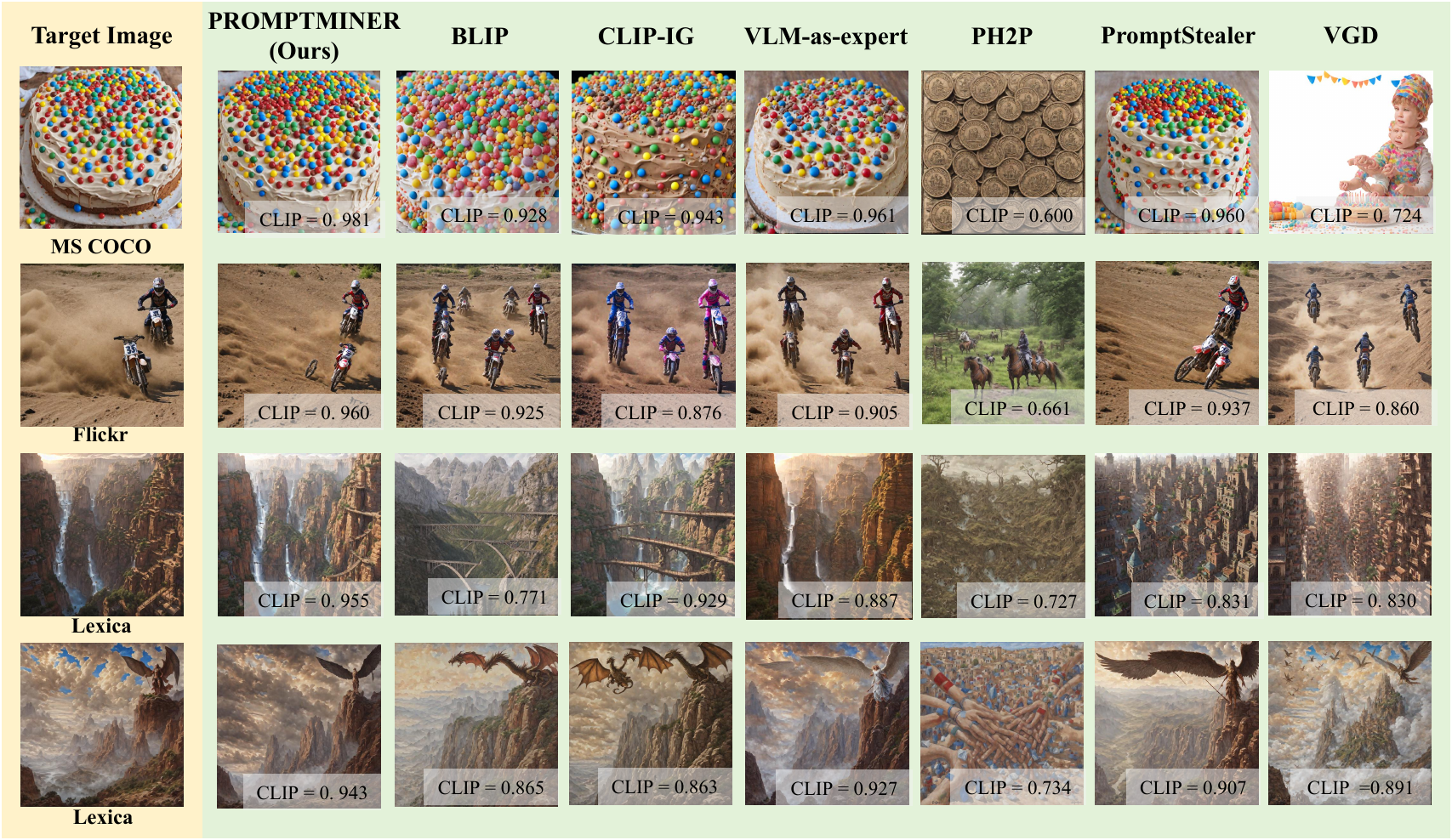} 
  \caption{Visualization of images generated by SDXL Turbo compared with target image.}
  \label{fig:img_sim_visual_sdxl}
\end{figure*}
\begin{figure*}[h] 
  \centering
  \includegraphics[width=1.0\linewidth]{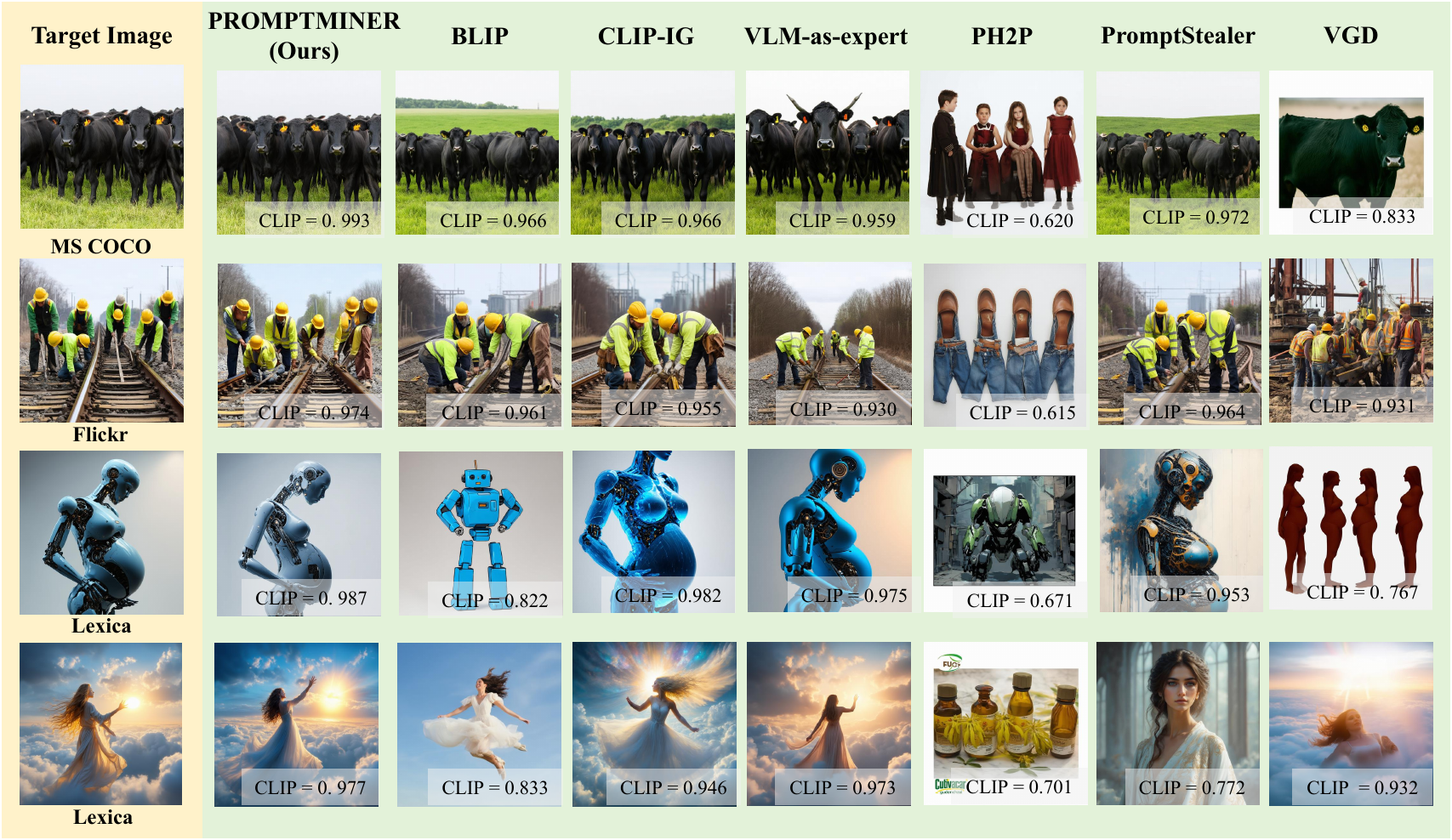} 
  \caption{Visualization of images generated by Stable Diffusion 3.5 Medium compared with target image.}
  \label{fig:img_sim_visual_sd35}
\end{figure*}
\begin{figure*}[h] 
  \centering
  \includegraphics[width=1.0\linewidth]{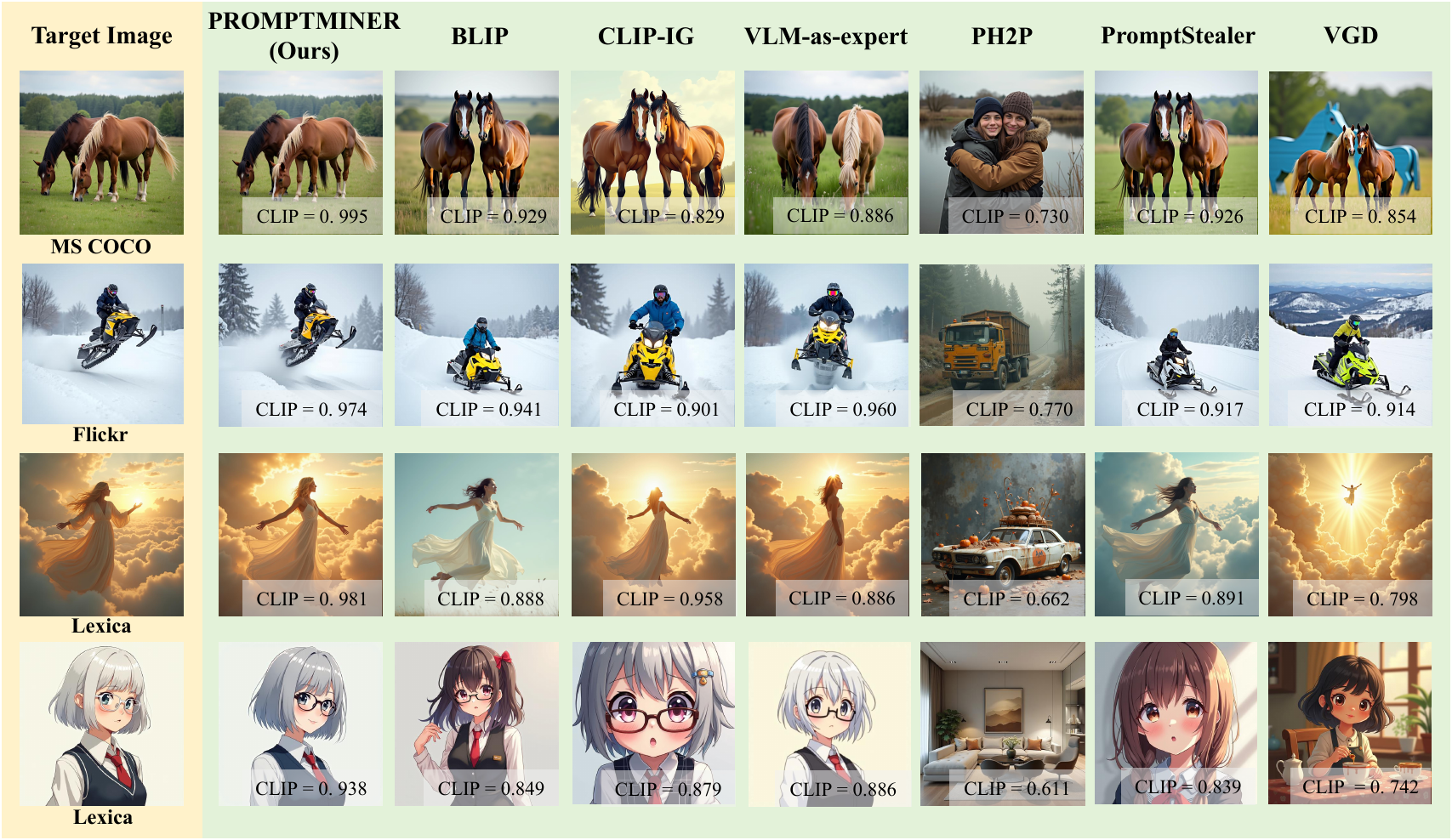} 
  \caption{Visualization of mages generated by FLUX.1 dev compared with target image.}
  \label{fig:img_sim_visual_flux}
\end{figure*}
\begin{figure*}[t] 
  \centering
  \includegraphics[width=1.0\linewidth]{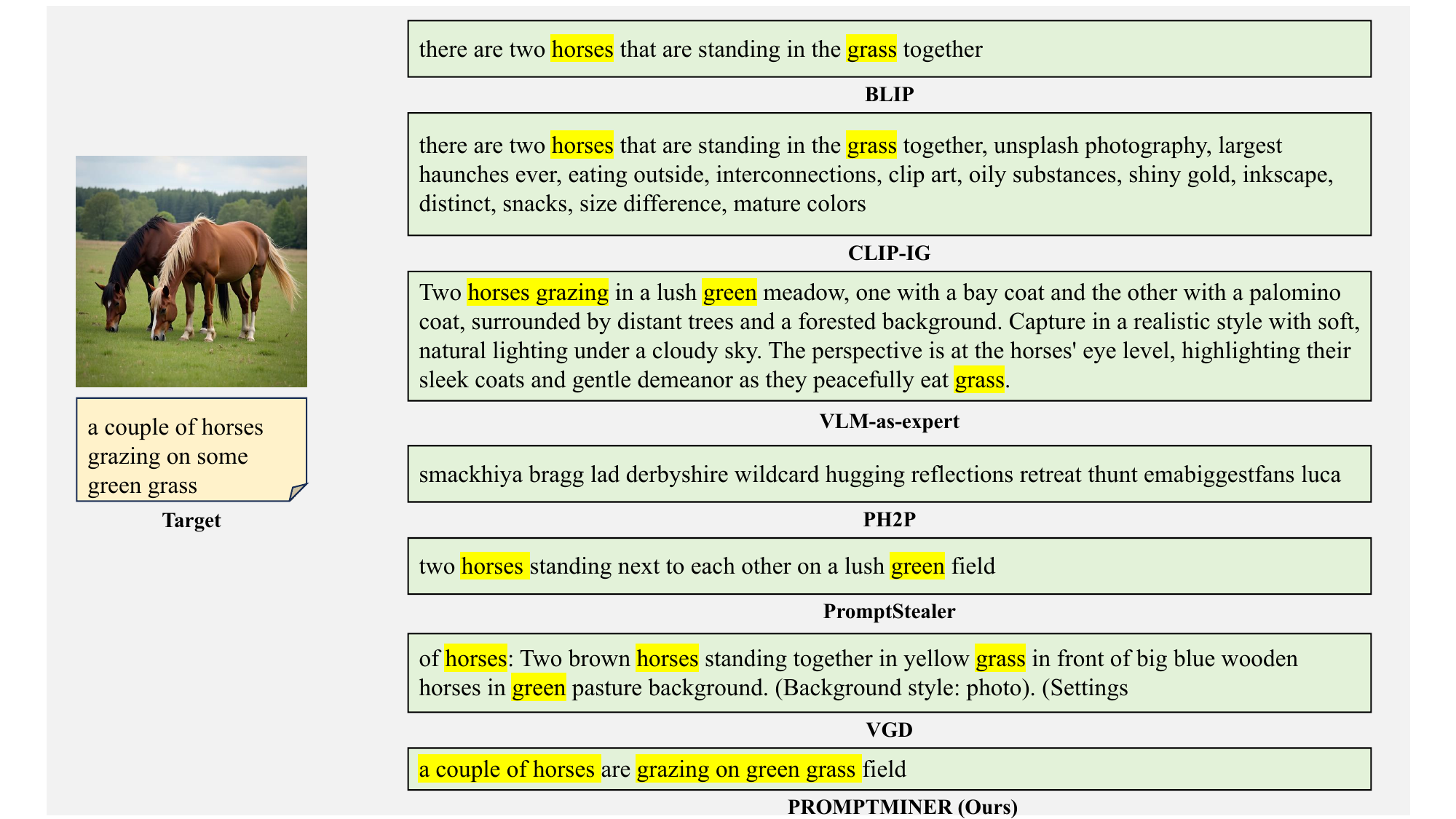} 
  \caption{Qualitative Results of stolen prompts compared with target prompt on MS COCO.}
  \label{fig:img_align_1}

\end{figure*}
\begin{figure*}[t] 
  \centering
  \includegraphics[width=1.0\linewidth]{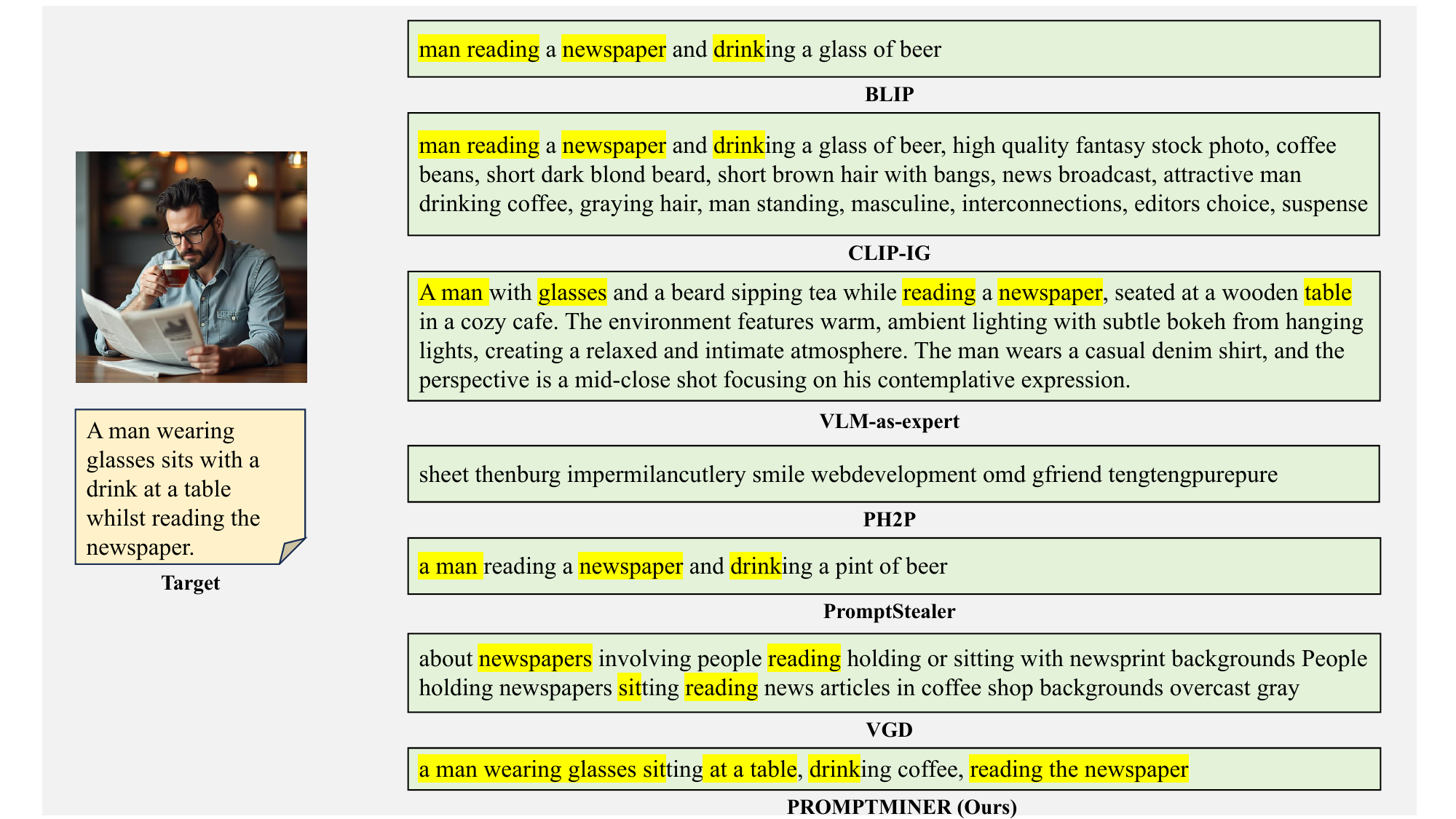} 
  \caption{Qualitative Results of stolen prompts compared with target prompt on Flickr.}
  \label{fig:img_align_2}
   
\end{figure*}
\begin{figure*}[t] 
  \centering
  \includegraphics[width=1.0\linewidth]{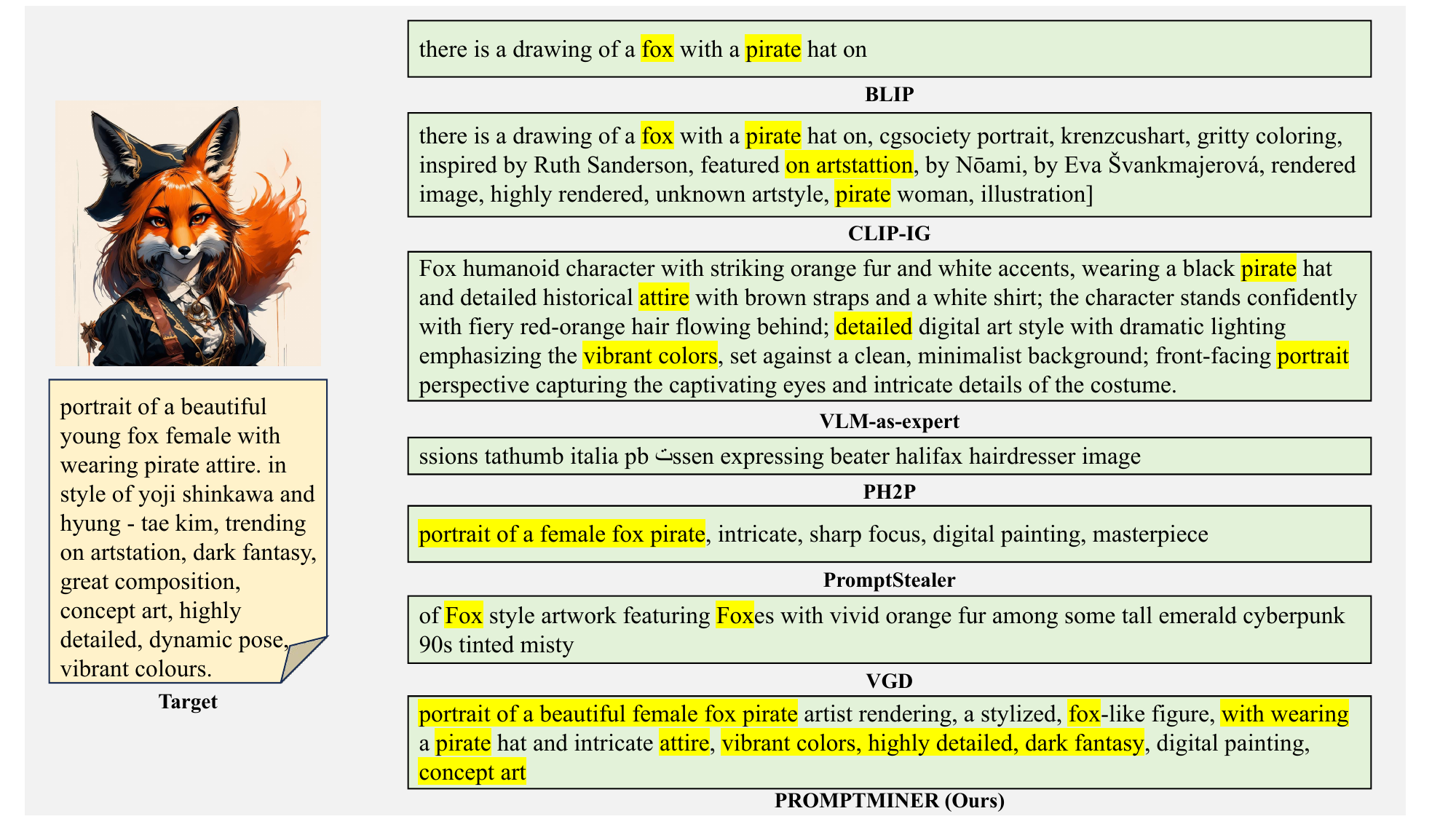} 
  \caption{Qualitative Results of stolen prompts compared with target prompt on Lexica.}
  \label{fig:img_align_3}
  
\end{figure*}

\end{document}